\documentclass[Afour,sageh,times]{sagej}

\usepackage{moreverb,url}
\usepackage{pdfpages}

\setcitestyle{aysep={,}}

\usepackage[colorlinks,bookmarksopen,bookmarksnumbered,citecolor=red,urlcolor=red]{hyperref}

\usepackage{subfigure}
\usepackage[switch]{lineno}
\newcommand\BibTeX{{\rmfamily B\kern-.05em \textsc{i\kern-.025em b}\kern-.08em
T\kern-.1667em\lower.7ex\hbox{E}\kern-.125emX}}

\def\volumeyear{2023}

\begin{document}

\runninghead{Thiago H Silva and Daniel Silver}

\title{Using Graph Neural Networks to Predict Local Culture}

\author{$^{\infty}$Thiago H Silva\affilnum{1} and $^{\infty}$Daniel Silver\affilnum{1,2}}

\affiliation{\affilnum{1}University of Toronto\\
\affilnum{2}Universidade Tecnologica Federal do Parana\\\vspace{.3cm} $\infty$ Equal contribution to this work.}

\corrauth{Thiago H Silva}

\email{thiagoh@utfpr.edu.br}

%\linenumbers
\begin{abstract}
Urban research has long recognized that neighbourhoods are dynamic and relational. However, lack of data, methodologies, and computer processing power have hampered a formal quantitative examination of neighbourhood relational dynamics. To make progress on this issue, this study proposes a graph neural network (GNN) approach that permits combining and evaluating multiple sources of information about internal characteristics of neighbourhoods, their past characteristics, and flows of groups among them, potentially providing greater expressive power in predictive models. By exploring a public large-scale dataset from Yelp, we show the potential of our approach for considering structural connectedness in predicting neighbourhood attributes, specifically to predict local culture. Results are promising from a substantive and methodologically point of view. Substantively, we find that either local area information (e.g. area demographics) or group profiles (tastes of Yelp reviewers) give the best results in predicting local culture, and they are nearly equivalent in all studied cases. Methodologically, exploring group profiles could be a helpful alternative where finding local information for specific areas is challenging, since they can be extracted automatically from many forms of online data. Thus, our approach could empower researchers and policy-makers to use a range of data sources when other local area information is lacking. 
\end{abstract}


\maketitle

\section{Introduction}

Since its early days, urban research has grappled with the challenges inherent in understanding the internal complexity of cities. On the one hand, cities are composed of distinct areas, such as neighbourhoods and districts, which have their own histories, identities, and impacts on local residents' and businesses' fates and fortunes. On the other, neighbourhoods are interconnected with one another, and movement among them occurs frequently and relatively freely. In terms elaborated by the early Chicago School, neighbourhoods are fully contextual. Their inner social organization is bound up with their pasts and futures, just as much as their histories are shaped by other communities around and connected to them; they are always making new versions of themselves, but the degree to which the new version resembles the previous one varies \citep{abbott2017department}.   

While early Chicago school researchers, among others, articulated the idea of full neighbourhood contextuality theoretically, incorporating it into research proved difficult. Local ethnographies could examine how the life of single neighbourhoods link out beyond themselves in space and time \citep{zorbaugh1983gold}. But a formal quantitative examination of neighbourhood relational dynamics was hampered by a lack of data, methodologies, and computer processing power. Flows (of people, finances, and more) between cities offered new research opportunities \citep{duncan2013metropolis} and opportunities to conceptualize cities as embedded in graphs/networks \citep{sassen2013global}. 

The advent of geo-referenced big data, such as location-based social networks, opened up corresponding possibilities to study neighbourhoods in a similarly relational way \citep{Phillips2019Social,Poorthuis2021Changing,Shelton2019Nature}. Joining these data sources with modern network-analytical methodologies, in turn, has inspired more precise conceptualizations of cities in terms of their structural connectedness, understood as the degree to which their neighbourhoods are linked by the movements of residents. The city itself is conceived as a graph where vertices are neighbourhoods and edges are movements between them by persons, businesses, ideas, money, and more \citep{phillips2021social,sampson2012great,papachristos2018,Graif2019Network,Candipan2021From,Daepp2021Small}. 

Many studies on the relational dynamics of neighbourhoods have been descriptive. These tend to identify clusters or sets of more or less interconnected neighbourhoods, often highlighting how parts of cities thought to be isolated are more related than one might think, or how some areas, while physically close, can be socially distant \citep{Shelton2015Social,Andris2019Threads}. Other research aims to incorporate neighbourhood graphs into predictive models, which often feature gentrification, crime, and disease \citep{papachristos2018,saxon2021local}, and typically use traditional methods, such as regression analysis. A few studies \citep{Gilling2021Predicting,Ilic2019Deep,Palafox2020Predicting,Reades2018Understanding,Thackway2021Building} have explored the use of various machine learning models, including standard neural networks, to predict future neighbourhood states based on past conditions. However, ours is the first to our knowledge to explore Graph Neural Networks (GNN) in this context. 

Our study seeks to make two contributions to this research literature. We:

\begin{enumerate}
    \item show the potential of graph neural networks (GNN) to predict local cultural characteristics, building models that incorporate a) local area information (e.g. FSA socio-economic data), b) neighbourhood mobility graphs, and c) group profiles (or any other sophisticated information) of individuals who move between neighbourhoods.

\item show the potential of Yelp as a data source for revealing structural connectedness in urban research. 
\end{enumerate}

Our GNN approach permits combining and evaluating multiple sources of information, providing greater expressive power to study neighbourhood evolution. Using our proposed model, we find, for instance: i) either local area information or group profiles give the best results in predicting local cultural dimensions (or scenes), and they are nearly equivalent in all studied cases; ii) while local area information is commonly used in similar tasks, group profiles have not been examined previously. This type of information could be a helpful alternative where it is hard to find local information for specific areas, since group profiles can be extracted automatically from many forms of online data, in ways that we illustrate in this study. Therefore, this modelling approach could empower researchers and policy-makers to use a range of data sources in situations where other local area information is lacking, such as in areas with poor census coverage or in the years between census data collection. 

The paper proceeds in the following sections. First, we discuss related work. Second, we review data and methods, featuring the GNN model we develop. Third, we present results. Our primary focus is on model evaluation, but here we also offer a tentative illustration of how GNN models can aid in substantive understanding. The fourth section discusses conclusions, limitations, and future directions.

\section{ Related Work}

\subsection{Relational approaches to neighbourhood research}

In the classical conception of the early Chicago School, ``Social facts are located. This means a focus on social relations and spatial ecology in synchronic analysis, as it means a similar focus on process in diachronic analysis" (\citep{abbott2017department} p. 197). Applied to neighbourhoods, this principle implies considering a neighbourhood as a vertex within an ``interactional field" in which every location and group is part of a ``whole network of intertwined processes" (\citep{abbott2017department} p. 200) whereby any given location is bound up both with what happens elsewhere and before. 

Yet while this insight informed much ethnographic and historical research through the 20th century, predominant forms of quantitative research have often faced challenges in incorporating it. ``Most sociological articles presume unrelated individuals, whether workers, firms, or associations" or neighbourhoods (\citep{abbott2017department} p. 197). These difficulties are largely due to limitations imposed by traditional data sources and methods. For example, in a national sample survey of 1,500 individuals, it is often not possible to examine these individuals' contexts and relations in a statistically meaningful way. Likewise, standard regression analysis treats each observation as independent. In the case of neighbourhoods, this often results in considering administrative units such as census tracts as if they were unrelated to others, even those nearby or with which they routinely exchange individuals, businesses, money, or ideas -- or proceeding as if an area's current state was not a single moment in an unfolding process of other preceding and following states.

Recent methodological advances and new fine-grained geo-referenced data sources have catalyzed a resurgence in quantitative studies of neighbourhoods that have given new empirical life to classical relational theories of neighbourhoods, both spatially and temporally (\cite{silver2022towards_part1} and \cite{knaap2019dynamics}). This work joins with rich bodies of multi-method research on local ``activity spaces" that spill across neighbourhood boundaries \citep{Cagney2020Urban}. Spatially, researchers have developed applications of graph methods to consider any given neighbourhood concerning others \citep{browning2017ecological,hipp2017neighborhoods,neal2012connected,papachristos2018,sampson2012great}, often defining the interconnectedness of neighbourhoods in terms of movements of people and firms between them \citep{phillips2021social}. Temporally, researchers experiment with various methods to situate any given neighbourhood concerning its past and future, such as Markov chains, spatial Markov chains and sequence analysis, or related methods \citep{10.1371/journal.pone.0245357}. At the cutting edge are efforts to combine both in spatio-temporal models \citep{delmelle2021neighborhood,SILVER2023104130,olson2021classification,dias2021neighborhood}. A number of spatial and contextual regression models enable the independence assumption to be relaxed and probed. 

Given the novelty of the data and rapid methodological innovations, much research on relational dimensions of neighbourhoods has been descriptive. This work has revealed important features of cities and communities. For example, \cite{Candipan2021From}  use geo-tagged Twitter data to build a ``segregated mobility index" to measure the degree to which neighbourhoods of various racial compositions are connected to others with similar or different compositions and use that index to characterize 50 US cities. \cite{Andris2019Threads} reveal neighbourhood graphs that span socio-economic differences via ties built by youth mentoring programs, while \cite{Shelton2015Social} use Twitter data to reveal the extent to which Louisville neighbourhoods are often thought to be separate and apart and are fluidly interconnected.  \cite{athey2021estimating} show that ``experienced isolation" based on neighbourhood mobility graphs is much lower than that revealed by studying residential isolation within neighbourhoods alone.  \cite{Daepp2021Small} uses consumer credit data and community detection methods to reveal clusters of strongly interconnected Massachusetts neighbourhoods.  \cite{Brazil2022Environmental} shows that poor neighbourhoods tend to be connected via mobility graphs to other neighbourhoods with higher pollution levels.  \cite{Poorthuis2021Changing} map how neighbourhood mobility graphs in Lexington, KY, evolved during gentrification. A significant body of related work aims to create novel metrics of neighbourhoods that do not rely on pre-determined administrative boundaries but rather activity spaces \citep{chen2019understanding}. Similarly, research examining temporal relations often primarily seeks to identify trajectories \citep{delmelle2021neighborhood}, showing, for example, that areas which might look similar at a given time point are on different evolutionary trajectories or, conversely, that areas which might look different at a given time point are moving in similar ways \citep{olson2021classification}.   

As research has generated new relational neighbourhood metrics, new opportunities have emerged to incorporate those metrics into predictive models. Gentrification, crime, and disease spread have been key outcomes of interest. For example, \cite{papachristos2018} show that criminal co-offending generates stable neighbourhood graphs and predicts spatial patterns in crime better than traditional spatial models.  \cite{levy2020triple} demonstrate that, across over 30,000 US neighbourhoods, those with greater levels of ``mobility based disadvantage" experience more homicides, controlling for several crucial variables;  \cite{levy2022neighborhood} build similar metrics out of daily mobility data in Wisconsin, San Francisco, and Seattle, and shows mobility disadvantage strongly predicts COVID-19 caseloads. \cite{saxon2021local} uses cellphone data to build neighbourhood mobility graphs in Chicago and shows that more strongly connected neighbourhoods experience lower levels of crime and greater educational and economic attainment; \cite{Graif2019Network} use commuting patterns and obtain similar results.  \cite{greenlee2019assessing} goes beyond assessing the impacts of mobility graphs (based on longitudinal household records) on distinct variables to show that a neighbourhood's multi-dimensional temporal trajectory strongly influences its residential mobility flows. 

\subsection{Neighbourhood research with machine learning models}

While many of the aforementioned studies utilize sophisticated modelling techniques, the use of graph neural networks (GNNs) to leverage mobility data and predict features of neighbourhoods has not yet been explored. Machine learning methods have begun to make their way into urban studies research in general and neighbourhood studies in particular. For example,  \cite{Gilling2021Predicting},  \cite{Reades2018Understanding}, and  \cite{Thackway2021Building} train various models to predict gentrification, including Decision Trees, XGBoost, Linear Regression, and Random Forests.  \cite{Noh2021Cafe} use an XGBoost model to predict commercial store opening locations in Seoul, Korea. Other studies have examined the potential of neural network models to predict land-use patterns ( \cite{Pijanowski2005Calibrating}). \cite{Ilic2019Deep} use a Siamese convolutional neural network to predict gentrification based on Google Street View images.  \cite{Palafox2020Predicting} use neural networks to predict gentrification in Mexico City and explore an Interpretability Method called LIME to understand which factors are driving the classification, while \cite{Papadomanolaki2019Detecting} combine fully convolutional networks and recurrent networks to detect urban stability and variability over time, using satellite data.

Recently, scholars have been exploring the power of GNNs for urban studies \citep{xiao2021predicting,yu2022recognition,guo2020optimized,hu2021urban,li2021prediction,xu2022application,li2021spatial}. For instance,  \cite{xiao2021predicting} explore GNNs to predict urban vibrancy, e.g. the presence of lively street life or the perception of social life in public spaces around subway stations.  \cite{yu2022recognition} propose using GNNs for the problem of drainage pattern recognition, showing that their approach outperforms other machine learning methods, including other deep learning approaches, e.g., convolutional neural networks.  \cite{guo2020optimized}, propose using GNNs to extract traffic networks' spatial and temporal features for improving traffic prediction (see also  \cite{li2021spatial}).  \cite{hu2021urban} use GNNs to predict urban functions at the road segment level, exploring traffic interaction data obtained from taxi trajectories. \cite{li2021prediction} propose an approach based on GNN to predict human activity intensity, which enables them to consider both spatial and social perspectives.  \cite{xu2022application} showed considerable performance improvement when using GNNs to classify areas into neighbourhood types (e.g. industrial, residential, commercial, or educational zones) based on visual cues.

This work indicates the relevance and promise of graphs to represent complex relationships in modelling urban and neighbourhood patterns. There are also several proposals for new variants of GNNs to deal with spatial data \citep{song2020spatial,zhu2021spatial}. For example,  \cite{zhu2021spatial} propose a model for spatial regression based on GNNs, which could be useful in various spatial regression analyses.  \cite{song2020spatial} propose a GNN for handling spatial-temporal data simultaneously. Even so, no previous study, to the best of our knowledge, has focused on applying GNNs in the context of predicting cultural dimensions of neighbourhoods from a mix of socio-demographic, mobility, and group profile information. In addition, not all GNN models support expressive graphs, i.e., those with vertex and edge attributes; hence another key contribution of our study is exploring the potential of such graphs in the context of neighbourhood research. Nevertheless, it must be stressed that the approach we evaluate here is one among many emergent machine learning techniques that may hold promise for neighbourhood research. Building a toolkit that compiles and makes available a range of methods is an important avenue for further research. 

\subsection{Local cultural dimensions}

While mobility graphs have begun to be utilized to predict outcomes such as gentrification, crime, and disease spread, they have not thus far been used to model local cultural dimensions. Even so, existing work points to the importance of this topic. Thus \cite{phillips2021social} suggest (but do not empirically test) the proposition that connected neighbourhoods should foster cultural diffusion, through the spread of ``tastes, fashion, values, attitudes, and cultural practices" \citep{harding2010reconsidering}. Cultural tastes spread both through personal ties, but also through the creation of and exposure to public displays of cultural dimensions in, for example, cafes, restaurants, parks, sidewalks, and various other amenities \citep{brown2017up,ghaziani2016there}. Such displays can include indications of acceptance of gay couples (e.g. rainbow flags), affirmations of transgressive or counter-cultural values (e.g. through body piercing studios), celebration of ethnic cultures (through themed restaurants or shops), or more  \citep{silver2016scenescapes}.  

These suggestions indicate that local cultural dimensions are the result of diverse and complex processes.  \cite{silver2022ChangingScene} synthesize multiple strands of research under the heading of the ``4 Ds": development, differentiation, diffusion, and defence. Development processes occur when local areas exhibit increases in income and education, which often generates a shift from more traditionalistic and communal cultural dimensions to more expressive and individualistic ones \citep{inglehart2007postmaterialist}. Differentiation is driven by competition among local businesses and services, pushing them to more specialized and unique offerings relative to elsewhere \citep{durkheim2014division}. Diffusion happens when new ideas are transmitted by people or communications technologies from one area to another \citep{tarde2010gabriel}. For example,  \cite{somashekhar2021can} finds that group profiles of new residents across gentrifying neighbourhoods vary greatly, spanning lifestyles such as ``Urban Achievers," ``Bohemian Mix," and ``American Dream," presumably contributing to distinct local cultures in their destinations despite similar socio-economic conditions. ``Defense" processes occur when established locals respond to newcomers with divergent tastes by reasserting their local cultural dimensions to maintain their current dimensions over time.  \cite{SILVER2023104130} develop quantitative models to identify when and where defence succeeds or fails to maintain existing local cultural dimensions, formalizing insights from several qualitative studies of single cases. 

 \cite{silver2022ChangingScene} gather evidence for each of these mechanisms, inspired by both classical and recent theories. However, while this and other research have articulated some mechanisms that may predict where and when some cultural styles are more or less likely to appear, the study of local cultural styles has not been formally integrated into neighbourhood graph research. Nor has the potential of machine learning in general or the expressive power GNNs in particular to combine multiple processes into a single model been explored. Instead, prior research on neighbourhood culture has examined various processes separately. Therefore, we build a model that combines  a) local area information (e.g. postal area socio-economic data), b) neighbourhood mobility graphs, and c) group profiles of individuals who move between neighbourhoods to predict future d) local public cultural dimensions or ``scenes"  \citep{silver2010scenes} embodied in venues such as cafes, restaurants, churches, night clubs, and more. More details on modelling and measurement are below. More details on cultural dimensions can be found in the \textit{Extra Information on Cultural Dimensions section in the Supplementary Material}.

\subsection{ Yelp as a source for studying neighbourhood variation and mobility graphs}

Many of the aforementioned studies have leveraged the analytical potential of social media to build neighbourhood mobility metrics \citep{9682710,FERREIRA2020240,senefonteSocInfo2020,8422972,Rodrigues17}. Geo-referenced tweets have been the most common source, along with mobile phone data and fine-grained administrative data. While Twitter holds great potential for mobility graph studies, it also has limits for some purposes, in that there is no guarantee that a tweet posted from a given location is about that location. In principle, many online activities, including location-based social networks, leave digital traces out of which mobility graphs could be constructed and which include information about places and activities that occur in the visited areas. For example,  \cite{Santala} explore Untappd check-ins to demonstrate the potential of using such type of data to uncover uses of urban areas that can change dynamically,  \cite{9682710} and \cite{FERREIRA2020240} have used Foursquare check-ins to trace tourists' group profiles internationally, and similar data could be used potentially at the neighbourhood level (see also \cite{arribas2016sociocultural} for another use of Foursquare data). 

While it has not featured prominently in neighbourhood graph studies, Yelp.com also has great potential to join mobility graph research with the study of local culture. Yelp is one of the most popular consumer review social media sites, with more than 150 million unique users providing millions of reviews in several different languages~\citep{ranard2016yelp}. Yelp has much to offer as a source of data for studying neighbourhood change. In particular, it contains information about (i) venues and (ii) their users. Venues are mainly businesses such as restaurants and coffee shops, but they also may include public spaces such as parks or hiking trails. Each venue is classified within a hierarchical list of categories and may contain user reviews containing evaluative assessments. The reviews are date stamped and correspond to a venue GPS location, enabling geospatial and temporal analysis. Furthermore, each user reviewer accumulates a portfolio of reviews and their locations.

Yelp has been utilized to study local neighbourhood culture. Such work tends to feature neighbourhood culture as revealed by local aggregations of review texts, characteristics of restaurants, or attributes of reviewers. For example,  \cite{Rahimi2018Geography},  \cite{HariniAutodetection},  and  \cite{olson2021reading} use computational text analysis to identify neighbourhood boundaries from review texts while creating metrics of local cultural similarity and difference;  \cite{Chen2021Analyzing} show that regional culture influences which cuisines tend to be combined in reviews.  \cite{rahimi2017using} home in on keywords around romance, dating, marriage, and family, finding that different parts of cities are more or less associated with different modes of romantic experience. \cite{zukin2017omnivore} examine how the sentiment of reviews varies across predominantly white or black New York neighbourhoods.  \cite{Glaeser2018Nowcasting}, by contrast, examines changing distributions of venue types (e.g., restaurants, grocery stores, bars, cafes), finding that changes in the business landscape are leading predictors of housing prices and that such dynamics can be observed in closer to real-time as compared to standard census sources (see also  \cite{dong2019predicting} and \cite{Rahimi2018Social}). Group profiles of reviewers (users) have been less frequently studied, but \cite{SILVER2023104130} suggest a method for doing so, and use that method to study how the local amenity mix of neighbourhoods is affected by the appearance of newcomers with tastes that do or do not diverge from those of the current residents.  

A key contribution of our study is to develop methods for combining the rich information contained in Yelp group profiles and to suggest directions for extending these methods to other sources. In particular, we combine four types of information. First, following  \cite{SILVER2023104130}, we build group taste profiles based on the sets of venues that individuals review. Reviewers who visit similar sets of venues are grouped together. Second, following  \cite{silver2016scenescapes}, we build profiles of neighbourhood public culture or ``scenes" based on their overall sets of venues (e.g. tattoo shops, night clubs, fashion houses, community centers), featuring the cultural orientations they support  (e.g. transgression, glamour, neighbourliness). Third, we use the location information associated with reviews to build neighbourhood graphs based on review co-location, where two neighbourhoods are considered connected when reviewers visit both of them. In addition, we use the date stamps on reviews to make these neighbourhood graphs dynamic, thereby combining spatial and temporal contextuality. The present study does not explicitly leverage review text, but we suggest directions for doing so in the conclusion. The GNN enables us to join and evaluate all this information into a single model alongside traditional census of population data.  

Building on the above review, we pursue this main research question: In predicting local cultural dimensions, what is the relative predictive power of i) area socio-economic data ii) group taste profiles and iii) mobility graphs?

\section{ Data and Method}

\subsection{ Overview}

\begin{figure*}[hbt!]  
\centering  
   \includegraphics[width=.99\textwidth]{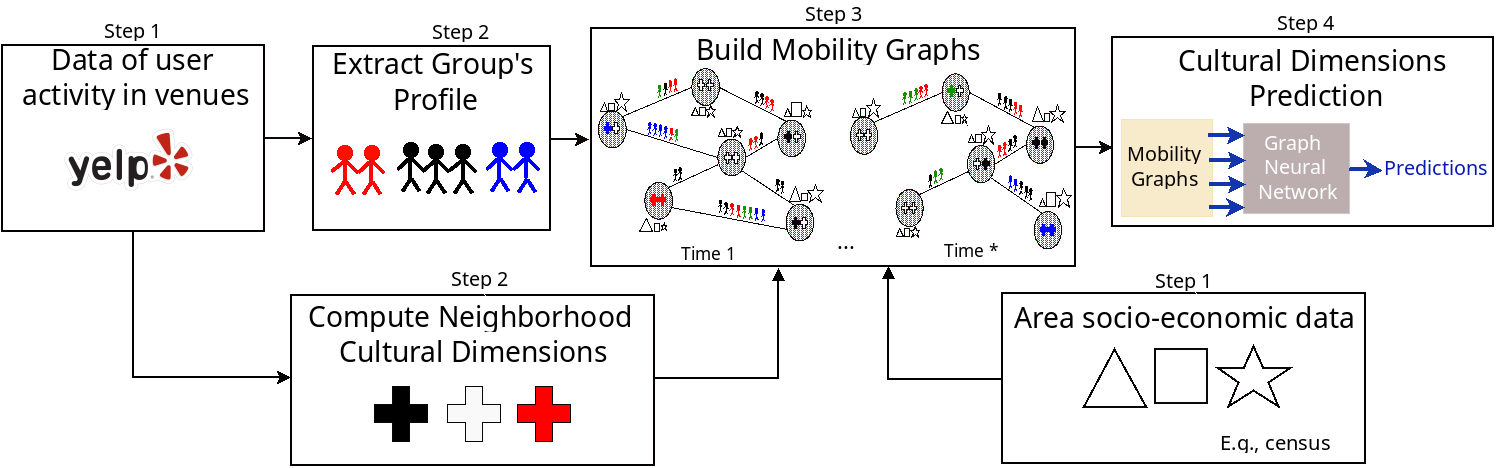}
\caption{Overall workflow of the study. Steps were numbered to indicate precedence order. Proceedings represented in Step 1 refer to data collection. Whereas those represented in Step 2 refer to information extraction on LBSN data. Steps 1 and 2 are necessary to build mobility graphs (Step 3), which, in turn, are explored in the cultural dimension predictions (Step 4). [Best in colour]}\label{fig:Overview}
\end{figure*}

Figure \ref{fig:Overview} illustrates the overall workflow of this study. We first collect Yelp data on user review activity -- note that other systems could be explored. Second, we extract for each user a group profile from the set of venues they visit, and define groups in terms of users sharing a similar group profile. Third, we use review co-location to construct mobility graphs for each year and each city in our dataset. Finally, we join the mobility and group profile information with census FSA socio-economic data in a GNN to predict neighbourhoods' future cultural dimensions -- neighbourhoods are represented in this study as FSAs. Details about data, measures, and methods are below. For the sake of simplicity, we explain our methodology for a specific city $p$. However, we repeat the same steps for all cities.

\subsection{Yelp dataset}

While our methodology could be applied to other similar datasets, we use an official dataset provided by Yelp \citep{yelpDataset}. This dataset includes a subset of Yelp venues, reviews, and user data from 2008 to 2018. Our study omits data prior to 2011 because before this time, the Yelp platform was still new, and its overall venue counts did not yet reliably correspond to counts from other sources \citep{SILVER2023104130}. We focus on the three most popular Canadian cities in this dataset (Calgary/Alberta, Montreal/Quebec, and Toronto/Ontario) because this makes it more efficient to join our various data sources. For each city, we have the count of venues, information about each venue's category (e.g. ``hair salon" or ``coffee shop"), and all of its reviews. Each review is also indexed to the reviewer who wrote it. We have a Venue ID, a User ID, and a location (lat/lon) from a venue associated with each review. Finally, in order to avoid including predominantly residential areas, we only consider areas (Foward Sortation Area - FSA\footnote{https://ised-isde.canada.ca/site/office-superintendent-bankruptcy/en/statistics-and-research/forward-sortation-area-fsa-and-north-american-industry-classification-naics-reports/forward-sortation-area-definition.}) with at least 30 unique venues. As several FSAs are residential, this filter left us with 14 in Montreal (22\% of all available), 26 in Calgary (74\%), and 38 in Toronto (38\%), all FSAs with commercial concentrations. Table \ref{tabOverData} summarizes our dataset containing the number of venues, reviews and users for all cities considered in this study.

\begin{table}[httb!]
\centering
\footnotesize
\caption{Overall statistics of the studied dataset.}
\begin{tabular}{ccccc}
           & \# Venues & \# Reviews & \# Users & \#  Categories\\ \hline
Calgary    & 7,736       & 97,650     & 34,645  & 774 \\
Montreal   & 6,602       & 155,966    & 58,660   & 639\\
Toronto    & 18,906      & 525,435    & 169,516 & 903
\end{tabular}
\label{tabOverData}
\end{table}

\subsection{Group Profile}

We identify group profiles in three steps, adapting for this application text analysis methods and following  \citep{SILVER2023104130}. 

First, we characterize each user by the categories of the venues they review. Users are considered as ``documents," and the categories of the items they review are treated as words. For example: $d_1$= \textit{\{Nightclub, Cafe, Sushi Restaurant, Mechanic\}}, where $d_1 \in D$  represents a document describing user1. 

Second, after applying standard pre-processing techniques (removing numbers, special characters, blank spaces, etc.), we identify latent topics in these cleaned documents using Latent Dirichlet Allocation (LDA) \citep{blei2003latent}. In this application, each user can be associated with topics; in this case, topics can be interpreted as the user's interests. To determine the number of topics to include in our analysis, we use the UMass topic coherence measure \citep{mimno2011optimizing} to evaluate 1 to 30 topics. This measure trains the topic model using the original corpus rather than relying on an external corpus, as in other coherence methods. We look at this distribution of coherence scores to identify a suggested number of topics, i.e., the highest score. We then map each document onto the topic space: $u = \{x_1, x_2,..., x_s\}$, where $u \in U$ is a feature vector representing a particular document, and  $x_i$ represents the probability that this user is associated with a certain topic among $s$ topics considered.

Third, we identify groups based on users' topics of interest. To do so, we find groups of users $C = \{c_1, \dots, c_k\}$ in the space represented by the feature vector $U$. We rely on the $k$-means algorithm, where $k$ is identified according to the silhouette heuristic (evaluating $k$ from 2 to 15). Each group identified is referred to as a group profile in this study.

\subsection{Area socio-economic data}

To measure socio-economic conditions in each area, we use data from the Canadian Census of Population (2016) and the National Household Survey (2011). We consider the following variables: percent with a BA or higher, average rent, percent classified as ``visible minorities," median income, percent age 20-34, percent who walk to work, and percent who work in arts, entertainment, and culture. We map the socio-economic data to the closest year of Yelp data. This set of variables captures neighbourhood socio-economic features, including ethno-racial composition, class, youth culture, housing affordability, walkability, and arts activity. While one might consider additional variables, these are similar to those used in past work on Canadian neighbourhood identity and change (e.g. \cite{silver2016scenescapes},  \cite{doering2021spatial}, and \cite{10.1371/journal.pone.0245357}). \textit{Census Files section, Figure S6, in the Supplementary Material} shows the differences between the census.

\subsection{Neighbourhood cultural dimensions}

The target of our prediction models is the local cultural dimensions or scenes for each FSA -- \textit{Extra Information on Cultural Dimensions section in the Supplementary Material} for more information. Following research on local ``scenescapes,'' we measure local scenes by aggregating the set of venue categories available in an area in terms of qualitative meanings they express. Detailed descriptions of the theoretical meaning of each dimension can be found in  \citep{silver2016scenescapes} and elsewhere with a focus on the Canadian context \citep{silver2017some}. To translate these concepts into measurements, for each venue category (e.g. tattoo parlour, fashion house, or farmers market), a team of trained coders assigned a score of 1-5 on a set of 15 cultural dimensions $t_i \in T =  \{t_1,t_2,...,t_{15}\}$, such as tradition, transgression, local authenticity, or glamour (described in \citep{silver2016scenescapes}). Each area (i.e. FSA) then receives a score for each of the 15 dimensions, calculated as a weighted average. More specifically, for a particular FSA, we have a vector $T_{FSA} = \{t^{FSA}_1,t^{FSA}_2,...,t^{FSA}_{15}\}$, where $t^{FSA}_i= \frac{1}{\omega}  \sum_{b=1}^{\omega} \left (  \frac{1}{m}\sum_{\phi=1}^{m}T^{b,\phi }_i \right ) $, with $\omega$ representing the number of unique venues in an FSA and $m$ the number of categories a venue has, and $T^{b,\phi}_i$ is the $i$-th element of the vector of cultural dimensions for a certain venue $b$ and one of its category $\phi$; thus, $t^{FSA}_i$ represents the average score of all venues found in the FSA for a specific cultural dimension considered -- regarding all categories for a venue with their average scores per cultural dimensions. For more details, see \textit{Extra Information on Cultural Dimensions section in the Supplementary Material}.

\begin{figure*}[htb!]  
\centering  
   \includegraphics[width=.99\textwidth]{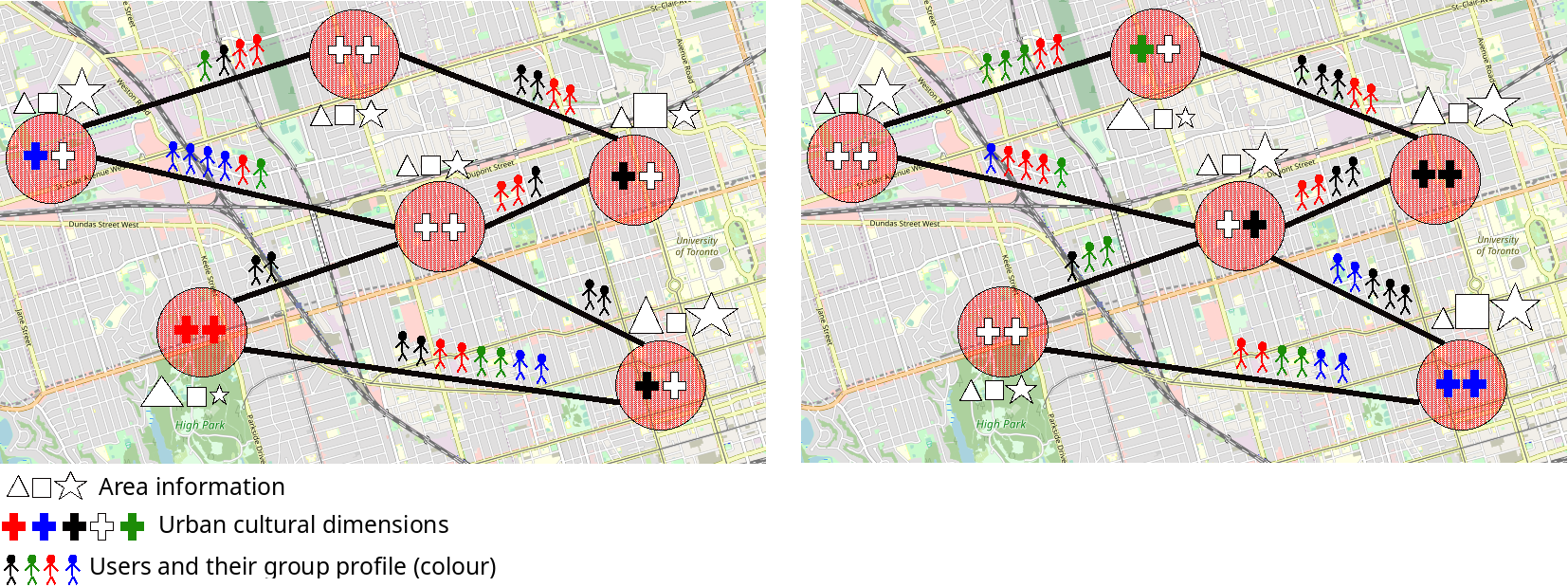}
\caption{Illustration of our mobility graph. Both figures represent the same city. On the left, we have data for year $y$, and, on the right, year $y+1$. For each year, we consider the same areas (vertices), shown as red circles. Note that areas can have attributes (i.e. the vertex attributes of the graphs), which are represented by the symbols triangle, square and star, where the size indicates the importance of the attribute. Crosses inside areas represent urban cultural dimensions. Edges can express the number of people that visited both areas (in all graphs) and the type of people that visited both areas, represented by the stick figure's colour (i.e. the edge attributes of the graph). [Best in colour]}\label{fig:mobilityNet}
\end{figure*}  

\subsection{Mobility Graph}\label{secmoblity}

Each city is characterized by an aggregated mobility graph per year describing users' visit patterns manifested in Yelp reviews. The mobility graph is an undirected weighted graph $G=(V,E)$ with a set $V=\{v_1,v_2,...,v_n \}$ of vertices corresponding to urban FSAs, and a set $E \subseteq V \times V$ of edges. If an edge $(v_i,v_j) \in E$, $e_{ij}$ represents two reviews made by the same user in the FSAs $v_i$ and $v_j$. All unique venues a user reviews in a certain time window will produce a complete graph pattern. I.e., if a user made reviews in $v_1$, $v2$, and $v3$ in a specific time window, then the graph referring to this period will have the edges  $(v_1,v_2)$, $(v_1,v_3)$, and $(v_2,v_3)$. A vertex $v$ is associated with vertex attributes/features $\mathbf{h}_v \in \mathbb{R}^{|T|+|D|}$, where $|T|$ is the number of features representing the urban cultural dimensions, and $|D|$ is the number of FSAs features in each vertex (socio-economic census data, in our study). An edge $e$ is associated edge features $\mathbf{h}_e \in \mathbb{R}^{|M|+|C|}$, where $|M|$ is one, corresponding to the mobility between two FSAs (i.e., edge weight), and $|C|$ corresponds to the number of group profiles identified for a specific city. Let $G^{(y)}_p=(V,E)$ represent a mobility graph, as described, considering data for a city $p \in \{Calgary,Montreal, Toronto\}$, in a specific year $y \in \{2011,2012, ..., 2018\}$.

Note that we can have distinct mobility graphs when considering the combination of $M$, $C$, and $D$. Table \ref{tabNamesNets} summarizes eight scenarios considered in this study according to this combination. As we can see in this table, all scenarios that consider census socio-economic data (i.e., vertex features $D$) have the term ``Area info'' in the name; when they take into account mobility weights (i.e., edge features $M$), they contain the term ``Mobility;'' and when they explore group profile (i.e., edge features $C$) the name has ``group profile.'' Note that scenario ``None'' is still a mobility graph, but without extra attributes on vertices and edges. Figure \ref{fig:mobilityNet} illustrates our mobility graph.

\begin{table}[]
\scriptsize
\caption{Scenarios exploring GNNs considered in this study. }
\begin{tabular}{p{2.7cm}|p{1.2cm}|p{1.5cm}|p{1.2cm}}
\textbf{Graph name}                   & \textbf{Vertex features D?} & \textbf{Edge features M?} & \textbf{Edge features C?} \\ \hline
\textbf{Area info + mobility + group profile} & X                         & X                         & X                         \\ \hline
\textbf{Area info + mobility}           & X                         & X                         &                           \\ \hline
\textbf{Area info + group profile}            & X                         &                           & X                         \\ \hline
\textbf{Area info}                      & X                         &                           &                           \\ \hline
\textbf{Mobility + group profile}             &                           & X                         & X                         \\ \hline
\textbf{Mobility}                       &                           & X                         &                           \\ \hline
\textbf{Group profile}                        &                           &                           & X                         \\ \hline
\textbf{None}                           &                           &                           &                          
\end{tabular}
\label{tabNamesNets}
\end{table}

\subsection{Prediction model}

\subsubsection{ Overview}

The success of deep learning in many domains is partially attributed to the rapidly developing computational resources (e.g., GPU), the availability of big training data, and the effectiveness of deep learning in extracting latent representations from Euclidean data (e.g., images, text, and videos). While machine learning techniques such as deep learning effectively capture hidden patterns of Euclidean data, there is an increasing number of applications representing data in graphs/networks. Graphs are a type of non-Euclidean data structure that models a set of objects (vertices) and their relationships (edges). They offer a high level of expressive power, suitable for studying systems across various areas, especially complex systems such as social networks, molecular interaction in chemistry, and, as we argue in this study, neighbourhood change. Graphs can have a variable amount of unordered vertices, i.e., they can be irregular, and vertices may have a different number of neighbours. Therefore, some critical operations (e.g., convolutions) are challenging to apply to the graph domain. 

For these and other reasons, there is growing interest in developing deep-learning techniques for graph data. Accordingly, newer models are emerging to handle the complexity of graph data \citep{wu2020comprehensive,zhou2020graph}. A Graph Neural Network (GNN) is a class of deep learning for processing data that can be defined as graphs/networks. Thus, GNNs improve other deep learning techniques that work with Euclidean data, offering the ability to learn complex patterns associated with connections between components \citep{wu2020comprehensive,zhou2020graph}. For instance, in social networks, a graph-based learning system can leverage the interactions between users and content to make highly precise recommendations. 

Our study investigates whether neighbourhood research can also benefit from this type of model. GNNs are especially promising in this setting, because we are attempting to model complex interactions, as illustrated in Figure \ref{fig:mobilityNet}.  More specifically, we show the power of GNNs for this type of application by examining the degree to which the interaction of groups of users with different types of urban areas influences the prediction of local cultural dimensions/scenes. \textit{Figure S4 in the Supplementary Material} illustrates this model visually.

Studying these sorts of complex relationships without graphs, while possible, is highly challenging. By using GNNs, however, our model can relatively straightforwardly handle complex mobility graphs that include (multiple) vertex and edge attributes. In this study, we consider eight different scenarios (i.e., graphs, as defined in Section ``Mobility Graph"). All graphs have information on urban cultural dimensions in the areas (vertices). We aim to predict the urban cultural dimensions for all areas under evaluation. 

\subsubsection{ Technical details}

The key idea of GNNs is to generate representations of vertices that depend on the graph's structure and any available feature information \citep{bookGraphLear}. Motivated by the success of Convolution Neural Networks (CNNs) in the computer vision discipline, many approaches re-define the idea of convolution for graphs \citep{wu2020comprehensive}. Thus, Graph Convolutional Network (GCN), outlined by  \cite{kipf2017semisupervised}, has proven to be one of the most effective and popular baselines of GNN architectures \citep{bookGraphLear}. In this study, we explore the architecture proposed by  \cite{li2020deepergcn}, called DeeperGCN, where the authors define, among other things, a novel generalized graph convolution (GENConv). This architecture enables training deep GCNs, overcoming a limitation in early versions of GCNs  \citep{li2020deepergcn}. In addition, GENConv allows learning on graphs with vertex attributes – as in all the architectures – and edge attributes beyond weights, whereas weights are the only edge attribute supported by the original GCN. Having a model that supports graphs with richer information is very important to tackle the neighbourhood change problem studied in this paper.

The GNN design examined in this study is outlined in \textit{Figure S5 in the Supplementary Material}. Edge and Vertex features of the input graph are encoded by a linear transformation. Multiple DeeperGCN layers then convolve the resulting encoded graph. One DeeperGCN layer consists of a batch normalization layer (LayerNorm), an activation function (ReLU), one dropout layer, one GENConv layer, and a residual connection layer – details can be found in \citep{li2020deepergcn}. We used five DeeperGCN layers. The evaluation of different configurations, e.g., with more/less layers, is out of the scope of this paper. After that, a final linear layer maps the encoded information to generate a prediction for all 15 cultural dimensions.

\subsubsection{ Evaluation}\label{secEval}

To evaluate each of the eight scenarios (graph types, see Section ``Mobility Graph") considered, we conducted different experiments by varying our data. The first step is to divide our dataset in two. If we envision predicting local cultural dimensions for year $y$, then we consider data for year $y$ as the Test set. Next, we use all information on previous years available in our dataset to compose the Training set. In other words, when aiming to predict the local cultural dimensions for  $G^{(2015)}_p=(V,E)$ for a city $p$, our Test set, we use the graphs  $G^{(2014)}_p=(V,E)$,  $G^{(2013)}_p=(V,E)$,  $G^{(2012)}_p=(V,E)$ and  $G^{(2011)}_p=(V,E)$ in the learning process of our model, our Training set, as we are studying data from 2011 to 2018. 

\begin{figure}[hbt!]  
\centering  
   \includegraphics[width=.47
   \textwidth]{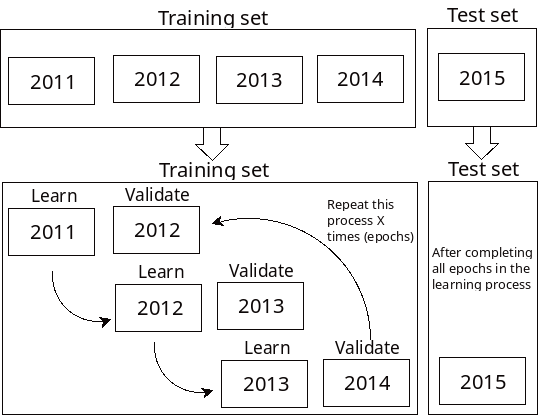}
\caption{ Diagram depicting the division of our dataset and procedures performed. This example assumes that we want to predict local cultural dimensions for the year 2015, thus, it becomes our test set. In this example, we train our model using all previous years: 2011 to 2014. One epoch (one complete cycle) in the training phase comprises learning and validation steps with all consecutive pairs of years, alternating learning and validation in each step, as depicted in the detailed part of the training set in the figure.}\label{fig:train}
\end{figure}

Figure \ref{fig:train} helps to illustrate our training and test process for the aforementioned scenario. As time is crucial in our dataset, we must consider it. We do so by adjusting our model by learning and validating on consecutive pairs of years of the Traning set. That is, we start by learning parameter values for a candidate model on data for 2011; then, we validate this model by trying to predict local cultural dimensions for 2012. We then calculate errors and adjust the parameters of our model. We repeat this process considering 2012 data for learning and 2013 for validating, and so on, for all data in the learning dataset. We repeat this process X times – in this study, we consider 10000 epochs.

\begin{figure*}[httt!]
\centering
 \subfigure[Calgary]
{\includegraphics[width=.44\textwidth]{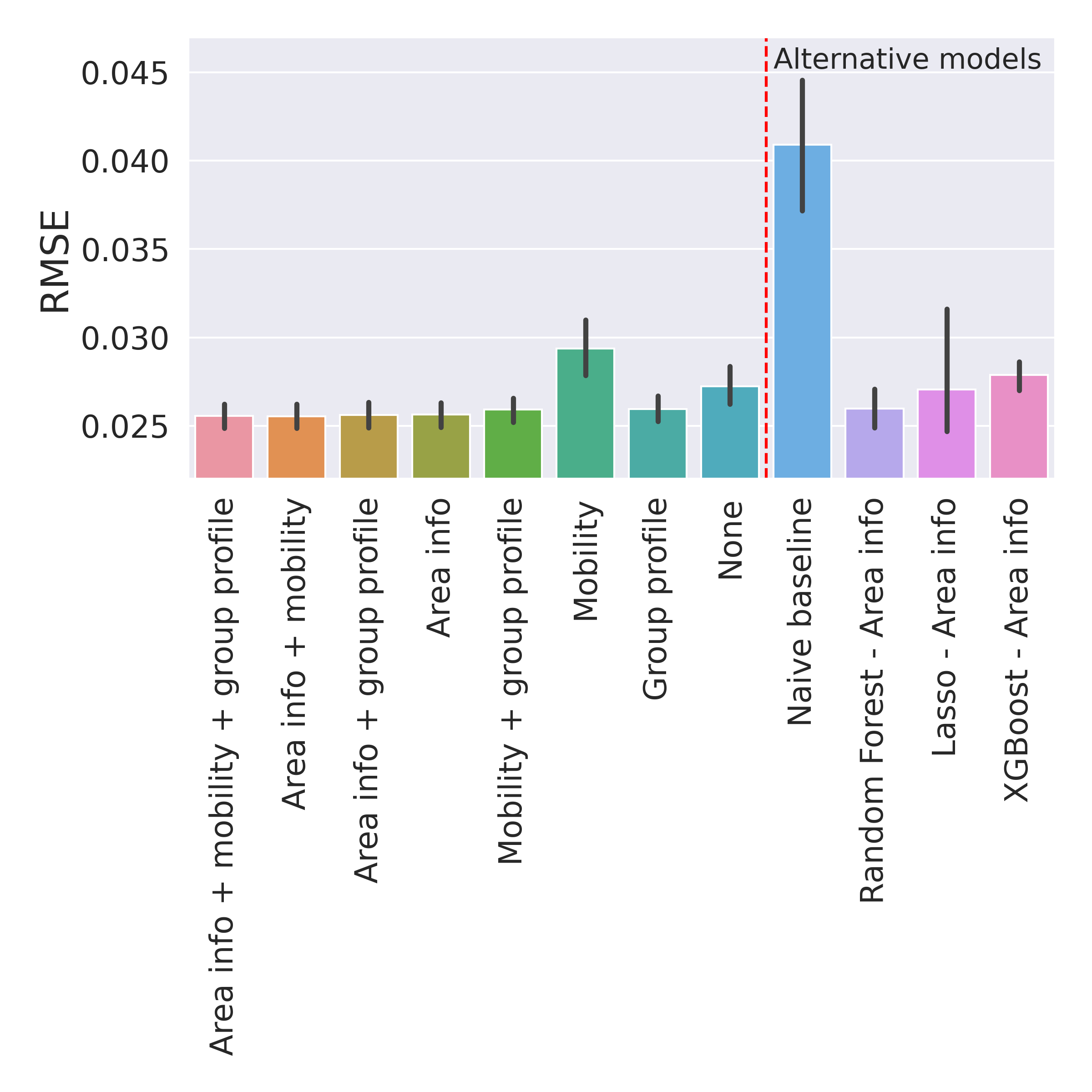} }
\subfigure[Montreal]
{\includegraphics[width=.44\textwidth]{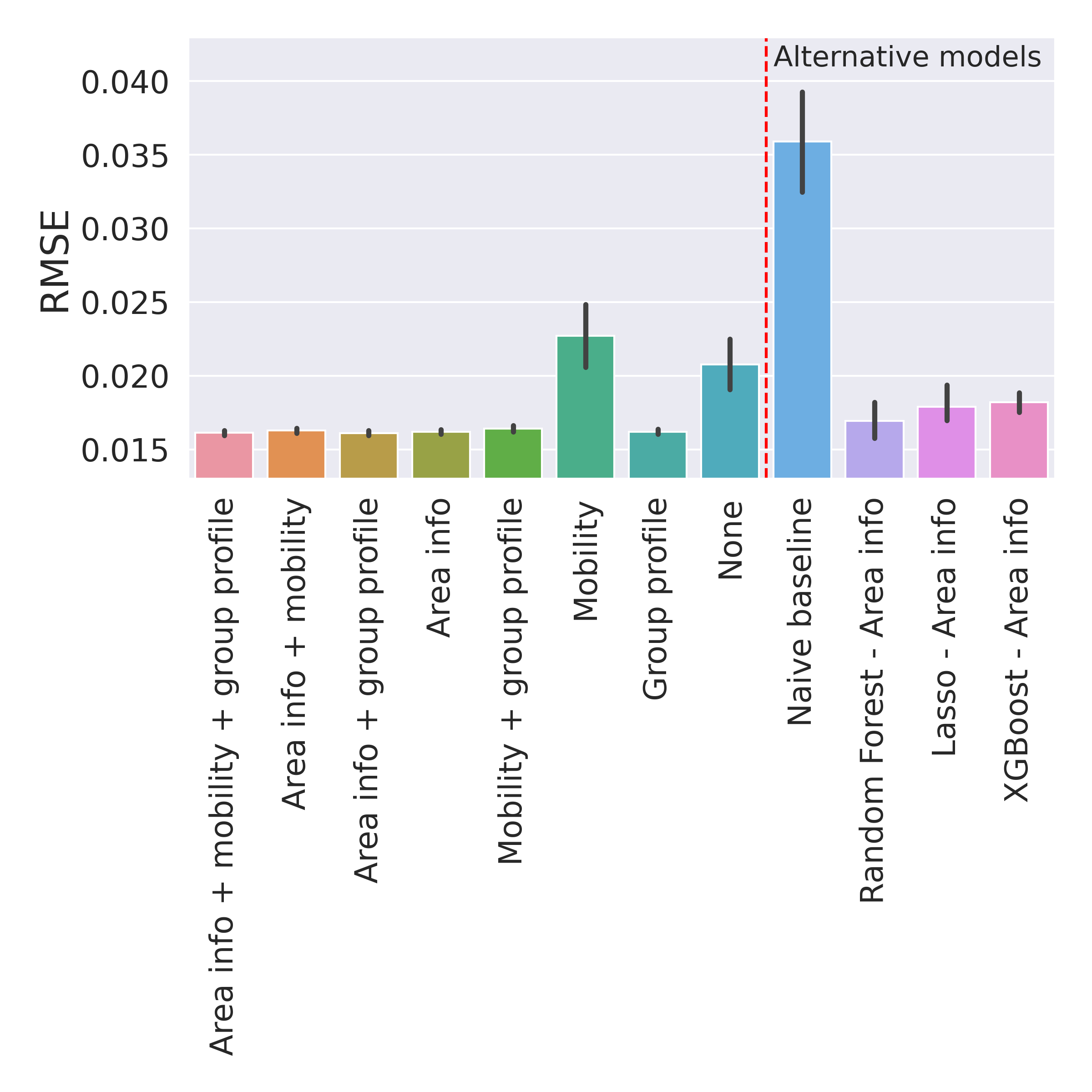}}
\subfigure[Toronto]
{\includegraphics[width=.44\textwidth]{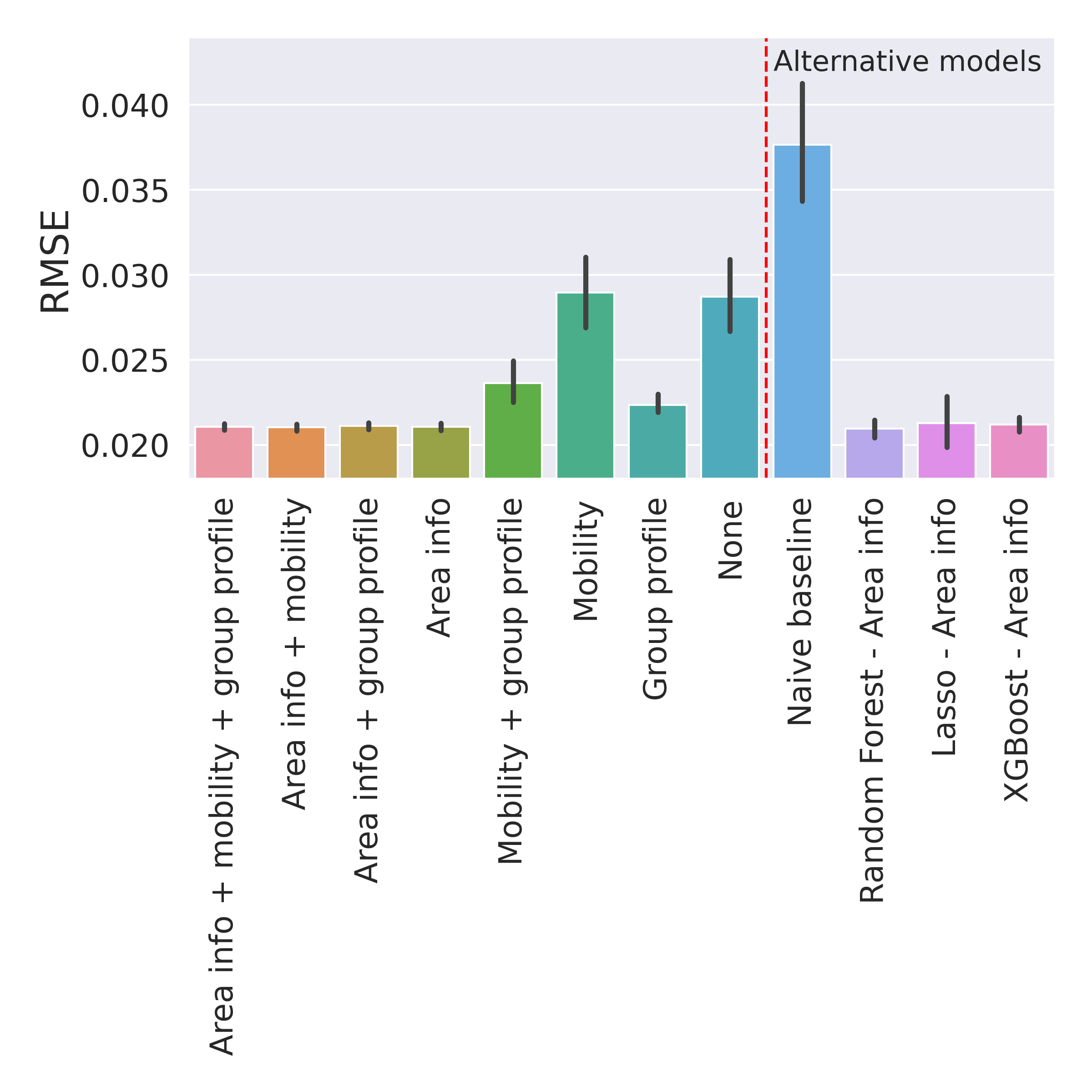}}
 \caption{Prediction errors (RMSE) for different scenarios (graphs) – the higher, the worse. Scenario None, a graph without census information, group profile and mobility information, presents the worst results for all cities. Note that we zoomed in on the y-axis to favour legibility. Note on group profiles: for Calgary and Montreal, we found five group profiles considering five topics; for Toronto, we found four group profiles considering seven topics. Recall that the GNN scenarios names are represented in Table \ref{tabNamesNets} -- for instance, ``area info'' is a scenario that only includes vertex features $D$. }
\label{figresults}
\end{figure*}

We predict local cultural dimensions for 2016, 2017, and 2018 for each scenario. In other words, these represent our Test set in a specific configuration, repeating experiments 25 times for each case. After that, for all experiments regarding a particular scenario in the city, we perform two types of analysis on the errors: i- aggregated view, computing the average RMSE (plus the confidence interval of 95\%) for all FSAs and cultural dimensions;  ii- errors by FSA, computing the average RMSE by each FSA for all dimensions -- here we also group FSAs by east and west regions to evaluate their distribution.

\subsubsection{Alternative models}

We also evaluate strategies not exploring graphs to contrast with the GNN approach. For these strategies, an FSA $f$ is associated with features $\mathbf{h}_{f} \in \mathbb{R}^{|T|+|D|}$, where $|T|$ is the number of features representing the urban cultural dimensions, and $|D|$ is the number of FSAs features. $H^{(y)}_p$ represents a set of FSAs and their features for a city $p \in \{Calgary,Montreal, Toronto\}$, in a specific year $y \in \{2011,2012, ..., 2018\}$. 

The evaluation process has similarities to the one presented in the previous section. The first step is dividing our dataset in two, where the year $y$ we want to predict is the Test set, and prior years are the Training set. We also predict local cultural dimensions for 2016, 2017, and 2018 for each alternative model. So, for instance, when predicting 2016, we consider $H^{(y)}_p$, where $y \in \{2011, ..., 2015\}$. We repeat experiments 25 times -- for all strategies that repetitions make sense. After that, we compute the average RMSE (plus the confidence interval of 95\%) for all results obtained.

The first strategy is called Naive Baseline. It considers the average cultural dimensions for each FSA of training data as a predictor. For instance, the predictions for the year 2016 is $\bar{H}_p = \frac{1}{|Y|}\sum_{y\in Y}^{}H^{(y)}_p$, where $Y = \{2011, ..., 2015\}$.

We also consider classical strategies: XGBoost \citep{chen2016xgboost}, Random Forests \citep{tan2016introduction}, and Lasso \citep{tibshirani1996regression}. We tuned each model's hyper-parameters, i.e., performing an exhaustive search over specified parameter values to find the best model in 5-fold cross-validation. Note that this process was not applied in the GNN strategies. i.e., it could be improved.

\section{Results}

Figure \ref{figresults} shows results for model evaluation. First, we look at the results based on GNNs -- scenarios described in Table \ref{tabNamesNets}. As expected, across all cities, not using any information other than the base structure of the graph -- the ``None" scenario --  produces one of the worst predictions of local cultural dimensions. The other situations we considered add content to the graph. In all cases, mobility information alone provides the worst predictions -- mobility cannot be reliably differentiated from ``None." This result implies that simply knowing about the degree of connectivity among neighbourhoods does not provide much information about their cultural characteristics. 

Adding more contextual information about areas or the groups that move among them improves the predictions. In general, area socio-economic information provides the best predictions. However, group profile information (taste profiles of groups associated with edges) gives statistically indistinguishable results in Calgary and Montreal. In Toronto, the difference is quite small (around .001 in terms of RMSE). Across all scenarios, there is no discernible improvement from combining information (e.g. mobility plus group profile does not perform better than group profile alone, area information plus other information does not perform better than area information alone). 

Now we concentrate on the analysis regarding the alternative models (not considering GNNs). First, we observe that the Naive baseline approach consistently produces the worst results -- around two times worse than the best results with GNNs. We also observe that there is always at least one result provided by Random Forest, Lasso, or XGBoost -- that only considers socio-economic data -- not distinguishable from the ``Area info'' scenario. This shows that if we are in possession of socio-economic census data, there is no evident advantage in using our studied GNNs approaches -- at least for the GNNs models we evaluated, without fine tuning as done for the alternative models.

These findings with regard to model evaluation are the central results of the paper. They show that in most cases, census socio-economic information is the best predictor of what the cultural dimensions in a neighbourhood will be. But, crucially, they also show that Yelp group profiles are generally as good. Thus, where census data might not be accessible -- between censuses or in places where socio-economic information might not be reliable -- if online review information is available, we may use them to learn about where the local culture of a place is likely to head. 

Nevertheless, while model evaluation is our primary goal here, we may also derive substantive insights from GNN models. We briefly illustrate a direction for doing so. These examples are offered in a tentative and experimental way to indicate possible directions for pursuing further social research questions. 

\begin{figure}[httt!]
\centering
\includegraphics[width=.49\textwidth]{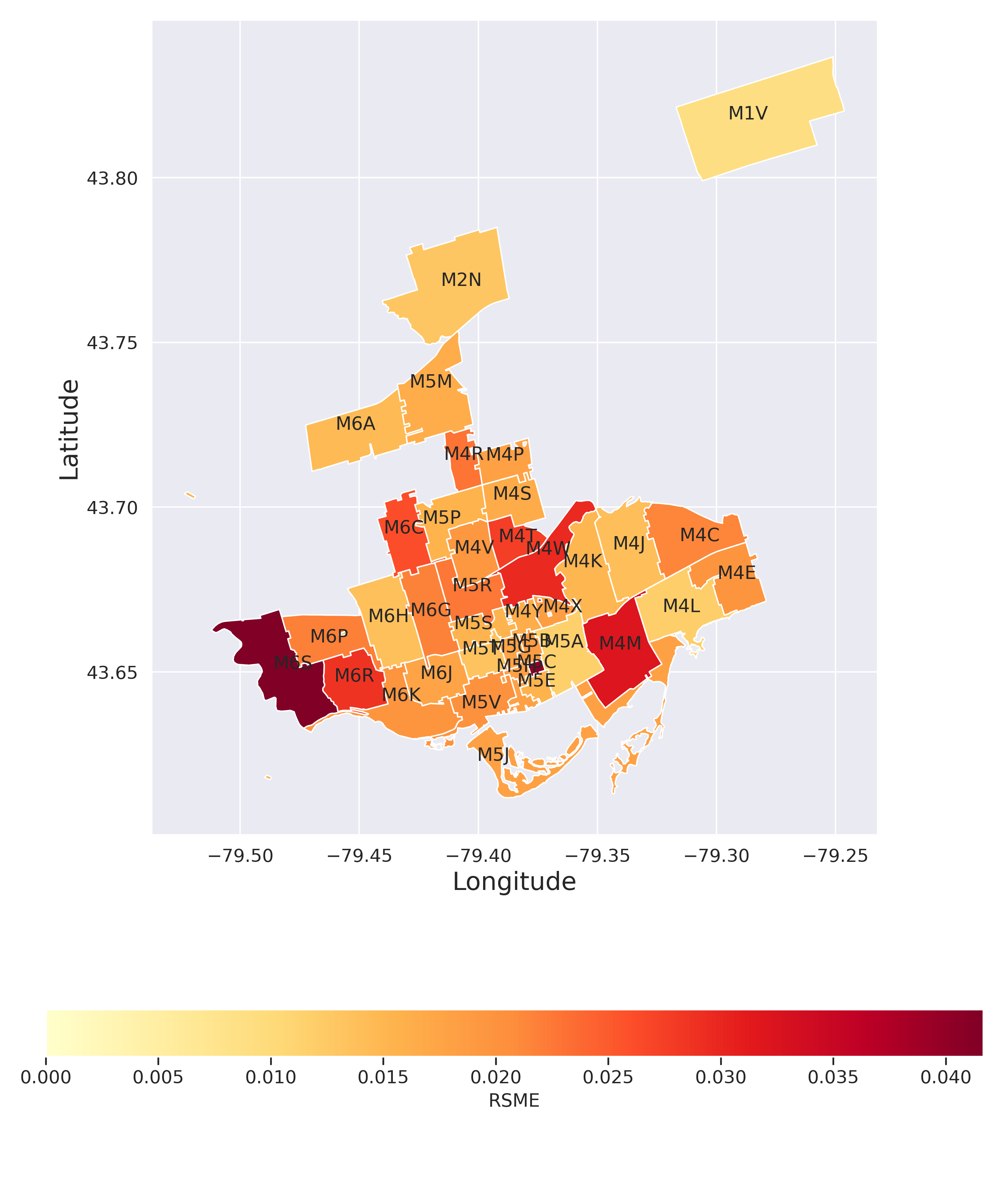} 
 \caption{Prediction errors (RMSE) for each FSA in Toronto.  Errors are presented in scenario "Area Info." [Best in colour]}
\label{figmaps}
\end{figure}

To do so, we may examine where the model's predictions are most accurate, i.e. variation in the error term. Here, we focus on Toronto for clarity -- see Figure \ref{figmaps}. What this map indicates is that ``error'' is not always or necessarily synonymous with ``bias.'' In fact, the areas of Toronto where the errors are higher seem to be those areas that are undergoing rapid transformations, which are less predictable from their pasts. For example, M6S and M6R (which encompass neighbourhoods such as Parkdale, Roncesvalles, Bloor West Village, and Runnymede) have experienced significant increases in new venues such as restaurants, bars, and arts establishments. Similarly, M4M includes Leslieville, where film studios have concentrated and new cultural dimensions have grown up around them. Also, errors tend to be higher in the west-end region -- see \textit{Figure S7 in the Supplementary Material}. While these results are not definitive, they suggest that we can use GNN findings as a starting point for deeper investigations of where and why the future is less strictly resembling the past \citep{silver2022towards_part1,fox2022towardsPart2,silver2022towards_part3,fox2022towards_PART4}\footnote{We explored an additional complementary analysis examining variation in prediction error for each cultural dimensions and found no statistically significant differences.}. 

\section{Discussion and Conclusion}

The primary contribution of this study is to demonstrate the potential of Graph Neural Networks for studying neighbourhoods in general and local culture in particular. Previous research has investigated features of neighbourhoods (such as socio-economic profiles), mobility graphs (often from Twitter and other location-based services), and the groups that move between them. Even in sophisticated prior efforts, these have rarely, if at all, been joined into a single model. The GNN approach we have proposed enables researchers to do so, in order to evaluate richer and more complex scenarios that would be hard (if possible at all) using other strategies.

Moreover, the experiments we conducted to test this method have substantive significance for urban and neighbourhood researchers. On the one hand, our findings indicate that, at least for predicting local cultural dimensions overall, traditional data sources such as national censuses remain critical. Socio-economic indicators from the census performed the best in all our experiments. On the other hand, our findings also suggest that researchers proceed with caution in treating mobility data alone as a strong predictor of neighbourhood cultural change. In our experiments, neighbourhood connectivity was the weakest predictor and, in some cases not statistically distinguishable from the empty graph. In the middle are the taste profiles of groups who move among neighbourhoods. 

Our findings suggest that certain groups of people could be a proxy for predicting urban cultural dimensions/scenes. While this should be taken with caution and further investigated, this result sheds light on avenues for exploring this type of feature in new ways. In situations where we do not have area attributes (our vertex features), we have the extra possibility of using group profiles to study neighbourhood change. For example, it provides a chance to observe changes occurring in the period between the US decennial census. At the same time, we found that for some cultural dimensions -- such as those that feature personal self-expression -- Yelp group profiles provide better predictions than census data. In such cases, we may perhaps gain greater insight and predictive power from the information in online review group profiles than from census socio-economic information. Nevertheless, for some dimensions -- such as ethically-oriented scenes -- our results point toward opportunities for more refined data to improve model performance.  

These findings open up opportunities for a number of future directions. For example, we may investigate whether certain groups (more than others) tend to produce outcomes, for instance, by causing a neighbourhood to take on cultural features that are conducive to their tastes, or sparking resistance \citep{SILVER2023104130}. In addition, we can create richer group profiles from Yelp data, and other internet data sources. For instance, the characterization of group profiles could be enriched by integrating information from users' reviews and more information about the venues they visit (such as price, ratings, ambiance, and the like). One could also create scene profiles for users, similar to scene profiles of areas. Other location-based social networks offer similar possibilities. In these and similar ways, a richer characterization of group profiles may match or surpass traditional area profiles from socio-economic data. We also can extend the analysis from single cities separately to sets of cities by combining them into a single graph. This would allow us to study the potential impact of inter-city and intra-city mobility graphs, as well as to pursue muti-level modelling frameworks. This would enhance the results we have presented regarding where the local cultural dimensions appear to be exhibiting the most or least variability. Finally, we could evaluate the creation of a single model to predict cultural dimensions for different cities in the same country instead of different models for that. This could be used for cities with fewer data. These are exciting possibilities that the GNN approach opens up. 

Despite the strong potential of GNN's for studying neighbourhood change revealed by this study, it is important to note some limitations. One of them is related to causal ordering. Based solely on our results, it is not possible to determine whether urban forms attract certain groups' profiles and, consequently, their change attracts different groups, or the other way around: new groups enter areas that generate pressure to change urban forms. This is a classic ``chicken and egg" problem, and addressing it is not a trivial task; often, the answer is ``both" \citep{silver2016scenescapes}. Researchers have historically pursued various regression-based frameworks for handling it, but exploring new avenues in a GNN context could prove illuminating. This would enhance certain existing theoretical and methodological frameworks that involve more non-linearities and feedback processes, such as the one proposed by \cite{SILVER2023104130}. Moreover, other strategies could be applied for identifying group profiles in addition to the one pursued here based on LDA. While any number of such strategies could be evaluated, our focus here is simply to demonstrate that one possible strategy to extract group profiles shows promise. Similarly, other cultural dimensions could be studied. Here, we featured the concept of scenes, but other representations could be considered with the necessary adaptations in the proposed framework. Also, while GNNs have great express power, they suffer from poor explainability, as other complex models.

While these are all exciting and promising directions, the present paper lays the ground to pursue them. We describe how to apply Graph Neural Network in predictive models of neighbourhood features in general and local cultural dimensions in particular, and demonstrate the potential of this approach. The technique shows considerable promise, both methodologically and substantively.

%%%%%%%%%%%%%%%%%%%%%%%%%%%%%%%%%%%%%%%%%%%%%%%%%%%%%%%%%

\section*{Declaration of conflicting interests}

The authors declared no potential conflicts of interest with respect to the research, authorship, and/or publication of this article.

\section*{Funding}

This research was partially supported by the SocialNet project (process 2023/00148-0 of São Paulo State Research Support Foundation - FAPESP) and by the National Council for Scientific and Technological Development - CNPq (processes 314603/2023-9 and 441444/2023-7).

\bibliographystyle{SageH}
\bibliography{main}

\begin{thebibliography}{93}
\providecommand{\natexlab}[1]{#1}
\providecommand{\url}[1]{\texttt{#1}}
\providecommand{\urlprefix}{URL }
\expandafter\ifx\csname urlstyle\endcsname\relax
  \providecommand{\doi}[1]{DOI:\discretionary{}{}{}#1}\else
  \providecommand{\doi}{DOI:\discretionary{}{}{}\begingroup \urlstyle{rm}\Url}\fi

\bibitem[{Abbott(2017)}]{abbott2017department}
Abbott A (2017) \emph{Department and discipline: Chicago sociology at one hundred}.
\newblock University of Chicago Press.

\bibitem[{Andris et~al.(2019)Andris, Liu, Mitchell, O\textquoteright{}Dwyer and Van~Cleve}]{Andris2019Threads}
Andris C, Liu X, Mitchell J, O\textquoteright{}Dwyer J and Van~Cleve J (2019) Threads across the urban fabric: Youth mentorship relationships as neighborhood bridges.
\newblock \emph{Journal of Urban Affairs} 43(1): 77--92.
\newblock \doi{10.1080/07352166.2019.1662726}.

\bibitem[{Arribas-Bel et~al.(2016)Arribas-Bel, Kourtit and Nijkamp}]{arribas2016sociocultural}
Arribas-Bel D, Kourtit K and Nijkamp P (2016) The sociocultural sources of urban buzz.
\newblock \emph{Environment and Planning C: Government and Policy} 34(1): 188--204.

\bibitem[{Athey et~al.(2021)Athey, Ferguson, Gentzkow and Schmidt}]{athey2021estimating}
Athey S, Ferguson B, Gentzkow M and Schmidt T (2021) Estimating experienced racial segregation in us cities using large-scale gps data.
\newblock \emph{Proceedings of the National Academy of Sciences} 118(46): e2026160118.

\bibitem[{Blei et~al.(2003)Blei, Ng and Jordan}]{blei2003latent}
Blei DM, Ng AY and Jordan MI (2003) Latent dirichlet allocation.
\newblock \emph{Journal of Machine Learning Research} 3(Jan): 993--1022.

\bibitem[{Brazil(2022)}]{Brazil2022Environmental}
Brazil N (2022) Environmental inequality in the neighborhood networks of urban mobility in {US} cities.
\newblock \emph{Proceedings of the National Academy of Sciences} 119(17).
\newblock \doi{10.1073/pnas.2117776119}.

\bibitem[{Brown-Saracino and Parker(2017)}]{brown2017up}
Brown-Saracino J and Parker JN (2017) ‘what is up with my sisters? where are you?’the origins and consequences of lesbian-friendly place reputations for lbq migrants.
\newblock \emph{Sexualities} 20(7): 835--874.

\bibitem[{Browning et~al.(2017)Browning, Calder, Soller, Jackson and Dirlam}]{browning2017ecological}
Browning CR, Calder CA, Soller B, Jackson AL and Dirlam J (2017) Ecological networks and neighborhood social organization.
\newblock \emph{American Journal of Sociology} 122(6): 1939--1988.

\bibitem[{Cagney et~al.(2020)Cagney, York~Cornwell, Goldman and Cai}]{Cagney2020Urban}
Cagney KA, York~Cornwell E, Goldman AW and Cai L (2020) Urban {Mobility} and {Activity} {Space}.
\newblock \emph{Annual Review of Sociology} 46(1): 623--648.
\newblock \doi{10.1146/annurev-soc-121919-054848}.

\bibitem[{Candipan et~al.(2021)Candipan, Phillips, Sampson and Small}]{Candipan2021From}
Candipan J, Phillips NE, Sampson RJ and Small M (2021) From residence to movement: The nature of racial segregation in everyday urban mobility.
\newblock \emph{Urban Studies} 58(15): 3095--3117.
\newblock \doi{10.1177/0042098020978965}.

\bibitem[{Chen et~al.(2019)Chen, Arribas-Bel and Singleton}]{chen2019understanding}
Chen M, Arribas-Bel D and Singleton A (2019) Understanding the dynamics of urban areas of interest through volunteered geographic information.
\newblock \emph{Journal of Geographical Systems} 21: 89--109.

\bibitem[{Chen and Guestrin(2016)}]{chen2016xgboost}
Chen T and Guestrin C (2016) Xgboost: A scalable tree boosting system.
\newblock In: \emph{Proceedings of the 22nd acm sigkdd international conference on knowledge discovery and data mining}. pp. 785--794.

\bibitem[{Chen and Park(2021)}]{Chen2021Analyzing}
Chen Z and Park J (2021) Analyzing {Cultural} {Assimilation} through the {Lens} of {Yelp} {Restaurant} {Reviews}.
\newblock In: \emph{2021 {IEEE} 8th {International} {Conference} on {Data} {Science} and {Advanced} {Analytics} ({DSAA})}. IEEE.
\newblock \doi{10.1109/dsaa53316.2021.9564170}.

\bibitem[{Daepp(2021)}]{Daepp2021Small}
Daepp MIG (2021) Small-area moving ratios and the spatial connectivity of neighborhoods: Insights from consumer credit data.
\newblock \emph{Environment and Planning B: Urban Analytics and City Science} 49(3): 1129--1146.
\newblock \doi{10.1177/23998083211051742}.

\bibitem[{Delmelle and Rey(2021)}]{delmelle2021neighborhood}
Delmelle EC and Rey SJ (2021) Neighborhood effects and neighborhood dynamics.
\newblock \emph{Geographical Analysis} 53(2): 167--169.

\bibitem[{Dias and Silver(2021)}]{dias2021neighborhood}
Dias F and Silver D (2021) Neighborhood dynamics with unharmonized longitudinal data.
\newblock \emph{Geographical Analysis} 53(2): 170--191.

\bibitem[{Doering et~al.(2021)Doering, Silver and Taylor}]{doering2021spatial}
Doering J, Silver D and Taylor Z (2021) The spatial articulation of urban political cleavages.
\newblock \emph{Urban Affairs Review} 57(4): 911--951.

\bibitem[{Dong et~al.(2019)Dong, Ratti and Zheng}]{dong2019predicting}
Dong L, Ratti C and Zheng S (2019) Predicting neighborhoods’ socioeconomic attributes using restaurant data.
\newblock \emph{Proceedings of the national academy of sciences} 116(31): 15447--15452.

\bibitem[{Duncan et~al.(2013)Duncan, Scott, Lieberson, Duncan and Winsborough}]{duncan2013metropolis}
Duncan OD, Scott WR, Lieberson S, Duncan BD and Winsborough HH (2013) \emph{Metropolis and region}.
\newblock Routledge.

\bibitem[{Durkheim(2014)}]{durkheim2014division}
Durkheim E (2014) \emph{The division of labor in society}.
\newblock Simon and Schuster.

\bibitem[{Ferreira et~al.(2020)Ferreira, Silva and Loureiro}]{FERREIRA2020240}
Ferreira AP, Silva TH and Loureiro AA (2020) Uncovering spatiotemporal and semantic aspects of tourists mobility using social sensing.
\newblock \emph{Computer Communications} 160: 240 -- 252.
\newblock \doi{https://doi.org/10.1016/j.comcom.2020.06.005}.

\bibitem[{Fox et~al.(2022{\natexlab{a}})Fox, Silver and Adler}]{fox2022towardsPart2}
Fox MS, Silver D and Adler P (2022{\natexlab{a}}) Towards a model of urban evolution: Part ii: Formal model.
\newblock \emph{Urban Science} 6(4): 88.

\bibitem[{Fox et~al.(2022{\natexlab{b}})Fox, Silver, Silva and Zhang}]{fox2022towards_PART4}
Fox MS, Silver D, Silva T and Zhang X (2022{\natexlab{b}}) Towards a model of urban evolution part iv: Evolutionary (formetic) distance—an interpretation of yelp review data.
\newblock \emph{Urban Science} 6(4): 86.

\bibitem[{Ghaziani(2016)}]{ghaziani2016there}
Ghaziani A (2016) \emph{There goes the gayborhood?}
\newblock Princeton University Press.

\bibitem[{Gilling et~al.(2021)Gilling, Mishra, Gibli and Hernandez}]{Gilling2021Predicting}
Gilling G, Mishra V, Gibli J and Hernandez D (2021) Predicting {Neighborhood} {Change} {Using} {Publicly} {Available} {Data} and {Machine} {Learning}.
\newblock \emph{SSRN Electronic Journal} \doi{10.2139/ssrn.3911354}.

\bibitem[{Glaeser et~al.(2018)Glaeser, Kim and Luca}]{Glaeser2018Nowcasting}
Glaeser EL, Kim H and Luca M (2018) Nowcasting {Gentrification}: Using {Yelp} {Data} to {Quantify} {Neighborhood} {Change}.
\newblock \emph{SSRN Electronic Journal} \doi{10.2139/ssrn.3123733}.

\bibitem[{Graif et~al.(2019)Graif, Freelin, Kuo, Wang, Li and Kifer}]{Graif2019Network}
Graif C, Freelin BN, Kuo YH, Wang H, Li Z and Kifer D (2019) Network {Spillovers} and {Neighborhood} {Crime}: A {Computational} {Statistics} {Analysis} of {Employment}-{Based} {Networks} of {Neighborhoods}.
\newblock \emph{Justice Quarterly} 38(2): 344--374.
\newblock \doi{10.1080/07418825.2019.1602160}.

\bibitem[{Greenlee(2019)}]{greenlee2019assessing}
Greenlee AJ (2019) Assessing the intersection of neighborhood change and residential mobility pathways for the chicago metropolitan area (2006--2015).
\newblock \emph{Housing Policy Debate} 29(1): 186--212.

\bibitem[{Guo et~al.(2020)Guo, Hu, Qian, Liu, Zhang, Sun, Gao and Yin}]{guo2020optimized}
Guo K, Hu Y, Qian Z, Liu H, Zhang K, Sun Y, Gao J and Yin B (2020) Optimized graph convolution recurrent neural network for traffic prediction.
\newblock \emph{IEEE Transactions on Intelligent Transportation Systems} 22(2): 1138--1149.

\bibitem[{Hamilton(2020)}]{bookGraphLear}
Hamilton WL (2020) Graph representation learning.
\newblock \emph{Synthesis Lectures on Artificial Intelligence and Machine Learning} 14(3): 1--159.

\bibitem[{Harding et~al.(2010)Harding, Lamont and Small}]{harding2010reconsidering}
Harding D, Lamont M and Small ML (2010) \emph{Reconsidering culture and poverty}, volume 629.
\newblock Sage.

\bibitem[{Hipp and Boessen(2017)}]{hipp2017neighborhoods}
Hipp JR and Boessen A (2017) Neighborhoods, social networks, and crime.
\newblock In: \emph{Challenging Criminological Theory}. Routledge, pp. 275--298.

\bibitem[{Hu et~al.(2021)Hu, Gao, Wu, Xu, Zhang, Cui and Gong}]{hu2021urban}
Hu S, Gao S, Wu L, Xu Y, Zhang Z, Cui H and Gong X (2021) Urban function classification at road segment level using taxi trajectory data: A graph convolutional neural network approach.
\newblock \emph{Computers, Environment and Urban Systems} 87: 101619.

\bibitem[{Ilic et~al.(2019)Ilic, Sawada and Zarzelli}]{Ilic2019Deep}
Ilic L, Sawada M and Zarzelli A (2019) Deep mapping gentrification in a large {Canadian} city using deep learning and {Google} {Street} {View}.
\newblock \emph{PLOS ONE} 14(3): e0212814.
\newblock \doi{10.1371/journal.pone.0212814}.

\bibitem[{Inglehart(2007)}]{inglehart2007postmaterialist}
Inglehart R (2007) {223 Postmaterialist Values and the Shift from Survival to Self‐Expression Values}.
\newblock \doi{10.1093/oxfordhb/9780199270125.003.0012}.
\newblock \urlprefix\url{https://doi.org/10.1093/oxfordhb/9780199270125.003.0012}.

\bibitem[{Kipf and Welling(2017)}]{kipf2017semisupervised}
Kipf TN and Welling M (2017) Semi-supervised classification with graph convolutional networks.
\newblock In: \emph{International Conference on Learning Representations}.

\bibitem[{Knaap et~al.(2019)Knaap, Wolf, Rey, Kang and Han}]{knaap2019dynamics}
Knaap E, Wolf L, Rey S, Kang W and Han S (2019) The dynamics of urban neighborhoods: a survey of approaches for modeling socio-spatial structure.
\newblock \emph{SocArXiv} .

\bibitem[{Levy et~al.(2020)Levy, Phillips and Sampson}]{levy2020triple}
Levy BL, Phillips NE and Sampson RJ (2020) Triple disadvantage: Neighborhood networks of everyday urban mobility and violence in us cities.
\newblock \emph{American Sociological Review} 85(6): 925--956.

\bibitem[{Levy et~al.(2022)Levy, Vachuska, Subramanian and Sampson}]{levy2022neighborhood}
Levy BL, Vachuska K, Subramanian S and Sampson RJ (2022) Neighborhood socioeconomic inequality based on everyday mobility predicts covid-19 infection in san francisco, seattle, and wisconsin.
\newblock \emph{Science Advances} 8(7): eabl3825.

\bibitem[{Li et~al.(2020)Li, Xiong, Thabet and Ghanem}]{li2020deepergcn}
Li G, Xiong C, Thabet A and Ghanem B (2020) Deepergcn: All you need to train deeper gcns.
\newblock \emph{arXiv preprint arXiv:2006.07739} .

\bibitem[{Li et~al.(2021)Li, Gao, Lu, Liu, Zhang and Tu}]{li2021prediction}
Li M, Gao S, Lu F, Liu K, Zhang H and Tu W (2021) Prediction of human activity intensity using the interactions in physical and social spaces through graph convolutional networks.
\newblock \emph{International Journal of Geographical Information Science} 35(12): 2489--2516.

\bibitem[{Li and Zhu(2021)}]{li2021spatial}
Li M and Zhu Z (2021) Spatial-temporal fusion graph neural networks for traffic flow forecasting.
\newblock In: \emph{Proceedings of the AAAI conference on artificial intelligence}, volume~35. pp. 4189--4196.

\bibitem[{Mimno et~al.(2011)Mimno, Wallach, Talley, Leenders and McCallum}]{mimno2011optimizing}
Mimno D, Wallach H, Talley E, Leenders M and McCallum A (2011) Optimizing semantic coherence in topic models.
\newblock In: \emph{Proc. of Empirical Methods in Natural Language Processing}. Edinburgh, UK, pp. 262--272.

\bibitem[{Neal(2012)}]{neal2012connected}
Neal ZP (2012) \emph{The connected city: How networks are shaping the modern metropolis}.
\newblock Routledge.

\bibitem[{Noh and Park(2021)}]{Noh2021Cafe}
Noh SC and Park JH (2021) Caf{\' e} and {Restaurant} under {My} {Home}: Predicting {Urban} {Commercialization} through {Machine} {Learning}.
\newblock \emph{Sustainability} 13(10): 5699.
\newblock \doi{10.3390/su13105699}.

\bibitem[{Olson et~al.(2021{\natexlab{a}})Olson, Calderon-Figueroa, Bidian, Silver and Sanner}]{olson2021reading}
Olson AW, Calderon-Figueroa F, Bidian O, Silver D and Sanner S (2021{\natexlab{a}}) Reading the city through its neighbourhoods: Deep text embeddings of yelp reviews as a basis for determining similarity and change.
\newblock \emph{Cities} 110: 103045.

\bibitem[{Olson et~al.(2021{\natexlab{b}})Olson, Zhang, Calderon-Figueroa, Yakubov, Sanner, Silver and Arribas-Bel}]{olson2021classification}
Olson AW, Zhang K, Calderon-Figueroa F, Yakubov R, Sanner S, Silver D and Arribas-Bel D (2021{\natexlab{b}}) Classification and regression via integer optimization for neighborhood change.
\newblock \emph{Geographical Analysis} 53(2): 192--212.

\bibitem[{Palafox and Ortiz-Monasterio(2020)}]{Palafox2020Predicting}
Palafox L and Ortiz-Monasterio P (2020) Predicting {Gentrification} in {Mexico} {City} using {Neural} {Networks}.
\newblock In: \emph{2020 {International} {Joint} {Conference} on {Neural} {Networks} ({IJCNN})}. IEEE.
\newblock \doi{10.1109/ijcnn48605.2020.9207685}.

\bibitem[{Papachristos and Bastomski(2018)}]{papachristos2018}
Papachristos AV and Bastomski S (2018) Connected in crime: The enduring effect of neighborhood networks on the spatial patterning of violence.
\newblock \emph{American Journal of Sociology} 124(2): 517--568.

\bibitem[{Papadomanolaki et~al.(2019)Papadomanolaki, Verma, Vakalopoulou, Gupta and Karantzalos}]{Papadomanolaki2019Detecting}
Papadomanolaki M, Verma S, Vakalopoulou M, Gupta S and Karantzalos K (2019) Detecting {Urban} {Changes} with {Recurrent} {Neural} {Networks} from {Multitemporal} {Sentinel}-2 {Data}.
\newblock In: \emph{IGARSS 2019 - 2019 {IEEE} {International} {Geoscience} and {Remote} {Sensing} {Symposium}}. IEEE.
\newblock \doi{10.1109/igarss.2019.8900330}.

\bibitem[{Phillips et~al.(2019)Phillips, Levy, Sampson, Small and Wang}]{Phillips2019Social}
Phillips NE, Levy BL, Sampson RJ, Small ML and Wang RQ (2019) The {Social} {Integration} of {American} {Cities}: Network {Measures} of {Connectedness} {Based} on {Everyday} {Mobility} {Across} {Neighborhoods}.
\newblock \emph{Sociological Methods \& Research} 50(3): 1110--1149.
\newblock \doi{10.1177/0049124119852386}.

\bibitem[{Phillips et~al.(2021)Phillips, Levy, Sampson, Small and Wang}]{phillips2021social}
Phillips NE, Levy BL, Sampson RJ, Small ML and Wang RQ (2021) The social integration of american cities: Network measures of connectedness based on everyday mobility across neighborhoods.
\newblock \emph{Sociological Methods \& Research} 50(3): 1110--1149.

\bibitem[{Pijanowski et~al.(2005)Pijanowski, Pithadia, Shellito and Alexandridis}]{Pijanowski2005Calibrating}
Pijanowski BC, Pithadia S, Shellito BA and Alexandridis K (2005) Calibrating a neural networkbased urban change model for two metropolitan areas of the {Upper} {Midwest} of the {United} {States}.
\newblock \emph{International Journal of Geographical Information Science} 19(2): 197--215.
\newblock \doi{10.1080/13658810410001713416}.

\bibitem[{Poorthuis et~al.(2021)Poorthuis, Shelton and Zook}]{Poorthuis2021Changing}
Poorthuis A, Shelton T and Zook M (2021) Changing neighborhoods, shifting connections: mapping relational geographies of gentrification using social media data.
\newblock \emph{Urban Geography} 43(7): 960--983.
\newblock \doi{10.1080/02723638.2021.1888016}.

\bibitem[{Rahimi et~al.(2017)Rahimi, Andris and Liu}]{rahimi2017using}
Rahimi S, Andris C and Liu X (2017) Using yelp to find romance in the city: A case of restaurants in four cities.
\newblock In: \emph{Proceedings of the 3rd ACM SIGSPATIAL Workshop on Smart Cities and Urban Analytics}. pp. 1--8.

\bibitem[{Rahimi and Bose(2018)}]{Rahimi2018Social}
Rahimi S and Bose M (2018) Social class and taste in the context of {US} cities.
\newblock In: \emph{Proceedings of the 2nd {ACM} {SIGSPATIAL} {Workshop} on {Geospatial} {Humanities}}. ACM.
\newblock \doi{10.1145/3282933.3282938}.

\bibitem[{Rahimi et~al.(2018)Rahimi, Mottahedi and Liu}]{Rahimi2018Geography}
Rahimi S, Mottahedi S and Liu X (2018) The {Geography} of {Taste}: Using {Yelp} to {Study} {Urban} {Culture}.
\newblock \emph{ISPRS International Journal of Geo-Information} 7(9): 376.
\newblock \doi{10.3390/ijgi7090376}.

\bibitem[{Ranard et~al.(2016)Ranard, Werner, Antanavicius, Schwartz, Smith, Meisel, Asch, Ungar and Merchant}]{ranard2016yelp}
Ranard BL, Werner RM, Antanavicius T, Schwartz HA, Smith RJ, Meisel ZF, Asch DA, Ungar LH and Merchant RM (2016) Yelp reviews of hospital care can supplement and inform traditional surveys of the patient experience of care.
\newblock \emph{Health Affairs} 35(4): 697--705.

\bibitem[{Reades et~al.(2018)Reades, De~Souza and Hubbard}]{Reades2018Understanding}
Reades J, De~Souza J and Hubbard P (2018) Understanding urban gentrification through machine learning.
\newblock \emph{Urban Studies} 56(5): 922--942.
\newblock \doi{10.1177/0042098018789054}.

\bibitem[{Rodrigues et~al.(2017)Rodrigues, Boukerch, Silva, Loureiro and Villas}]{Rodrigues17}
Rodrigues D, Boukerch A, Silva TH, Loureiro A and Villas L (2017) Smaframework: Urban data integration framework for mobility analysis in smart cities.
\newblock In: \emph{Proc.\ of the 20th ACM International Conference on Modeling, Analysis and Simulation of Wireless and Mobile Systems}. Miami, USA.

\bibitem[{Sampson(2012)}]{sampson2012great}
Sampson RJ (2012) \emph{Great American city: Chicago and the enduring neighborhood effect}.
\newblock University of Chicago Press.

\bibitem[{Santala et~al.(2020)Santala, Costa, Gomes-Jr, Gadda and Silva}]{Santala}
Santala V, Costa G, Gomes-Jr L, Gadda T and Silva TH (2020) On the potential of social media data in urban planning: Findings from the beer street in curitiba, brazil.
\newblock \emph{Planning Practice \& Research} 35(5): 510--525.
\newblock \doi{10.1080/02697459.2020.1767394}.

\bibitem[{Santos et~al.(2018)Santos, Rodrigues, Silva, Loureiro, Pazzi and Villas}]{8422972}
Santos FA, Rodrigues DO, Silva TH, Loureiro AAF, Pazzi RW and Villas LA (2018) Context-aware vehicle route recommendation platform: Exploring open and crowdsourced data.
\newblock In: \emph{2018 IEEE International Conference on Communications (ICC)}. pp. 1--7.
\newblock \doi{10.1109/ICC.2018.8422972}.

\bibitem[{Sassen(2013)}]{sassen2013global}
Sassen S (2013) \emph{The global city: New york, london, tokyo}.
\newblock Princeton University Press.

\bibitem[{Saxon(2021)}]{saxon2021local}
Saxon J (2021) The local structures of human mobility in chicago.
\newblock \emph{Environment and Planning B: Urban Analytics and City Science} 48(7): 1806--1821.

\bibitem[{Senefonte et~al.(2020)Senefonte, Frizzo, Delgado, Luders, Silver and Silva}]{senefonteSocInfo2020}
Senefonte H, Frizzo G, Delgado M, Luders R, Silver D and Silva T (2020) {Regional Influences on Tourists Mobility Through the Lens of Social Sensing}.
\newblock In: \emph{Proc.\ of the International Conference on Social Informatics (SocInfo'20)}. Pisa, Italy.

\bibitem[{Senefonte et~al.(2022)Senefonte, Delgado, Lüders and Silva}]{9682710}
Senefonte HCM, Delgado MR, Lüders R and Silva TH (2022) Predictour: Predicting mobility patterns of tourists based on social media user’s profiles.
\newblock \emph{IEEE Access} 10: 9257--9270.
\newblock \doi{10.1109/ACCESS.2022.3143503}.

\bibitem[{Shelton and Poorthuis(2019)}]{Shelton2019Nature}
Shelton T and Poorthuis A (2019) The {Nature} of {Neighborhoods}: Using {Big} {Data} to {Rethink} the {Geographies} of {Atlanta}\textquoteright{}s {Neighborhood} {Planning} {Unit} {System}.
\newblock \emph{Annals of the American Association of Geographers} 109(5): 1341--1361.
\newblock \doi{10.1080/24694452.2019.1571895}.

\bibitem[{Shelton et~al.(2015)Shelton, Poorthuis and Zook}]{Shelton2015Social}
Shelton T, Poorthuis A and Zook M (2015) Social {Media} and the {City}: Rethinking {Urban} {Socio}-{Spatial} {Inequality} {Using} {User}-{Generated} {Geographic} {Information}.
\newblock \emph{SSRN Electronic Journal} \doi{10.2139/ssrn.2571757}.

\bibitem[{Silver(2017)}]{silver2017some}
Silver D (2017) \emph{Some Scenes of Urban Life}.
\newblock SAGE.

\bibitem[{Silver et~al.(2022{\natexlab{a}})Silver, Adler and Fox}]{silver2022towards_part1}
Silver D, Adler P and Fox MS (2022{\natexlab{a}}) Towards a model of urban evolution—part i: Context.
\newblock \emph{Urban Science} 6(4): 87.

\bibitem[{Silver et~al.(2010)Silver, Clark and Navarro~Yanez}]{silver2010scenes}
Silver D, Clark TN and Navarro~Yanez CJ (2010) Scenes: Social context in an age of contingency.
\newblock \emph{Social forces} 88(5): 2293--2324.

\bibitem[{Silver et~al.(2022{\natexlab{b}})Silver, Fox and Adler}]{silver2022towards_part3}
Silver D, Fox MS and Adler P (2022{\natexlab{b}}) Towards a model of urban evolution—part iii: Rules of evolution.
\newblock \emph{Urban Science} 6(4): 89.

\bibitem[{Silver et~al.(2022{\natexlab{c}})Silver, Silva and Adler}]{silver2022ChangingScene}
Silver D, Silva T and Adler P (2022{\natexlab{c}}) Changing the scene: applying four models of social evolution to the scenescape.
\newblock \emph{Journal of Wuhan University Technology (Philosophy and Social Sciences Edition)} 75: 49--65.
\newblock \doi{10.14086/j.cnki.wujss.2022.05.005}.

\bibitem[{Silver and Silva(2021)}]{10.1371/journal.pone.0245357}
Silver D and Silva TH (2021) A markov model of urban evolution: Neighbourhood change as a complex process.
\newblock \emph{PLOS ONE} 16(1): 1--29.
\newblock \doi{10.1371/journal.pone.0245357}.

\bibitem[{Silver and Clark(2016)}]{silver2016scenescapes}
Silver DA and Clark TN (2016) \emph{Scenescapes: How qualities of place shape social life}.
\newblock University of Chicago Press.

\bibitem[{Silver and Silva(2023)}]{SILVER2023104130}
Silver DA and Silva TH (2023) Complex causal structures of neighbourhood change: Evidence from a functionalist model and yelp data.
\newblock \emph{Cities} 133: 104130.
\newblock \doi{https://doi.org/10.1016/j.cities.2022.104130}.

\bibitem[{Somashekhar(2021)}]{somashekhar2021can}
Somashekhar M (2021) Can we bring culture into the large-scale study of gentrification? assessing the possibilities using geodemographic marketing data.
\newblock \emph{Urban Affairs Review} 57(5): 1312--1342.

\bibitem[{Song et~al.(2020)Song, Lin, Guo and Wan}]{song2020spatial}
Song C, Lin Y, Guo S and Wan H (2020) Spatial-temporal synchronous graph convolutional networks: A new framework for spatial-temporal network data forecasting.
\newblock In: \emph{Proceedings of the AAAI conference on artificial intelligence}, volume~34. pp. 914--921.

\bibitem[{Suresh and Locascio(2015)}]{HariniAutodetection}
Suresh H and Locascio N (2015) Autodetection and classification of hidden cultural city districts from yelp reviews.
\newblock \emph{arXiv preprint arXiv:1501.02527} .

\bibitem[{Tan et~al.(2016)Tan, Steinbach and Kumar}]{tan2016introduction}
Tan PN, Steinbach M and Kumar V (2016) \emph{Introduction to data mining}.
\newblock Pearson Education India.

\bibitem[{Tarde(2010)}]{tarde2010gabriel}
Tarde G (2010) \emph{Gabriel Tarde on communication and social influence: Selected papers}.
\newblock University of Chicago Press.

\bibitem[{Thackway et~al.(2023)Thackway, Ng, Lee and Pettit}]{Thackway2021Building}
Thackway W, Ng M, Lee CL and Pettit C (2023) Building a predictive machine learning model of gentrification in sydney.
\newblock \emph{Cities} 134: 104192.

\bibitem[{Tibshirani(1996)}]{tibshirani1996regression}
Tibshirani R (1996) Regression shrinkage and selection via the lasso.
\newblock \emph{Journal of the Royal Statistical Society Series B: Statistical Methodology} 58(1): 267--288.

\bibitem[{Wu et~al.(2020)Wu, Pan, Chen, Long, Zhang and Philip}]{wu2020comprehensive}
Wu Z, Pan S, Chen F, Long G, Zhang C and Philip SY (2020) A comprehensive survey on graph neural networks.
\newblock \emph{IEEE Transactions on Neural Networks and Learning Systems} 32(1): 4--24.

\bibitem[{Xiao et~al.(2021)Xiao, Lo, Zhou, Liu and Yang}]{xiao2021predicting}
Xiao L, Lo S, Zhou J, Liu J and Yang L (2021) Predicting vibrancy of metro station areas considering spatial relationships through graph convolutional neural networks: The case of shenzhen, china.
\newblock \emph{Environment and Planning B: Urban Analytics and City Science} 48(8): 2363--2384.

\bibitem[{Xu et~al.(2022)Xu, Jin, Chen, Xie, Hu and Xie}]{xu2022application}
Xu Y, Jin S, Chen Z, Xie X, Hu S and Xie Z (2022) Application of a graph convolutional network with visual and semantic features to classify urban scenes.
\newblock \emph{International Journal of Geographical Information Science} 36(10): 2009--2034.

\bibitem[{Yelp(2020)}]{yelpDataset}
Yelp (2020) Yelp dataset.
\newblock \url{https://www.yelp.com/dataset}.

\bibitem[{Yu et~al.(2022)Yu, Ai, Yang, Huang and Yuan}]{yu2022recognition}
Yu H, Ai T, Yang M, Huang L and Yuan J (2022) A recognition method for drainage patterns using a graph convolutional network.
\newblock \emph{International Journal of Applied Earth Observation and Geoinformation} 107: 102696.

\bibitem[{Zhou et~al.(2020)Zhou, Cui, Hu, Zhang, Yang, Liu, Wang, Li and Sun}]{zhou2020graph}
Zhou J, Cui G, Hu S, Zhang Z, Yang C, Liu Z, Wang L, Li C and Sun M (2020) Graph neural networks: A review of methods and applications.
\newblock \emph{AI open} 1: 57--81.

\bibitem[{Zhu et~al.(2021)Zhu, Liu, Yao and Fischer}]{zhu2021spatial}
Zhu D, Liu Y, Yao X and Fischer MM (2021) Spatial regression graph convolutional neural networks: A deep learning paradigm for spatial multivariate distributions.
\newblock \emph{GeoInformatica} : 1--32.

\bibitem[{Zorbaugh(1983)}]{zorbaugh1983gold}
Zorbaugh HW (1983) \emph{The gold coast and the slum: A sociological study of Chicago's near north side}.
\newblock University of Chicago Press.

\bibitem[{Zukin et~al.(2017)Zukin, Lindeman and Hurson}]{zukin2017omnivore}
Zukin S, Lindeman S and Hurson L (2017) The omnivore’s neighborhood? online restaurant reviews, race, and gentrification.
\newblock \emph{Journal of Consumer Culture} 17(3): 459--479.

\end{thebibliography}



\section*{Supplementary material}

# Supplementary Material

**Extra Information on Cultural Dimensions**

Similar scores have been constructed by teams working in the USA [1], Canada [2], France [3], Spain [4], China [5], Poland [6], and South Korea [7]. Mateos-Mora et al. [4] develop more formal evaluations of these measurements, while prior work has demonstrated substantial face, construct, and hypothesis validity (Silver and Clark [8], ch. 8). Silver, Clark, and Navarro [9] discuss various assumptions of this measurement approach.

We extend this past work by developing methods to apply qualitative coding to new datasets. Past work featured amenities data collected during 2005-2009 from online yellow pages directories (yellowpages.ca, superpages.com) and censuses of business (Zip Code Business Patterns in the US, Canadian Business Patterns in Canada). We map qualitative coding weights from those past datasets and their amenity categories to Yelp venue categories. This enables us to update analyses and carry them forward into the future without necessarily having to undertake the labour-intensive process of qualitative coding.

To accomplish this mapping, we performed a semi-automatic process to ease the burden of this task: the idea is to perform a transfer learning approach, importing the coding from previously validated sources. Categories on Yelp belong to a hierarchy. For example, the Italian category is under Restaurant. Thus, for each category on Yelp, we retrieve all categories that it might have above it in the hierarchy; for Italian, we consider the "sentence": Italian - Restaurant. We do the same for all categories of the systems for which we already have validated coding. Then, for each Yelp category (i.e., its "sentence"), we explore Sentence Transformers, which enables finding semantic similarities between sentences [10]. For each Yelp category, we get the seven best candidates provided by this technique among the sources for which we already have qualitative coding. A human evaluates these candidates to select the best match (if any). If there is no good match, which happened in a very small number of cases, the protocol was to look manually for a better match across all sources.

We validated our procedure by correlating the cultural dimensions produced with Yelp (after our process) and the previous coding with other sources for all FSA areas in Toronto. We chose Toronto because it was the only city where we have data for Yelp and the other sources. More specifically, we considered data for year $y$ in the Yelp dataset and each cultural dimension $t_i$. We then correlate, using Pearson and Spearman correlation, the vectors $\text{Yelp}_y = t_i^{FSA1}, t_i^{FSA2}, \ldots, t_i^{FSAq}$ and $\text{Source} = t_i^{FSA1}, t_i^{FSA2}, \ldots, t_i^{FSAn}$, where $n$ represents all FSA for Toronto, $t_i$ represents a specific cultural dimension, and *Source* represent the original codings we have (they do not vary on time). Nearly all correlations are positive, and most are high for all cultural dimensions $T$ and years considered in our Yelp dataset (2011 to 2018), indicating that our approach captured the information well, considering that we are mapping data from different sources and times onto one another -- see Figure S4. This matching procedure resulted in a tool to help transfer learning cultural dimensions coding to other sources, where further information can be found [11, 12].

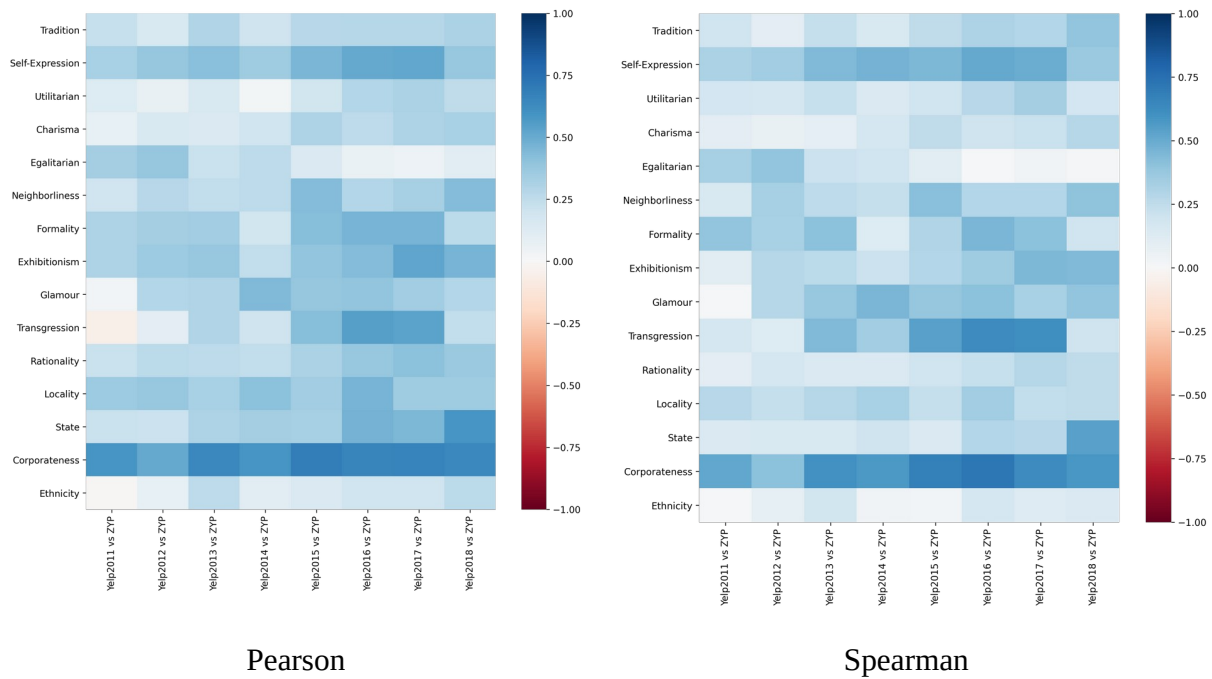

                      Pearson                                  Spearman

**Figure S1.** Correlation of the cultural dimensions produced with Yelp (after our process) and the previous coding with other sources for all FSA areas in Toronto. [Best in colour]

We illustrate the profile of cultural dimensions we are predicting by simply displaying the average and standard deviation values across Toronto: Charisma=3.12 (+-0.05), Corporateness=3.07 (+-0.04), Egalitarian=3.06 (+-0.02), Ethnicity=3.19 (+-0.04), Exhibitionism=3.02 (+-0.02), Formality=3.03 (+-0.03), Glamour=3.11 (+-0.05), Locality=3.06 (+-0.04), Neighborliness=3.15 (+-0.04), Rationality=2.78 (+-0.05), Self-Expression=3.11 (+-0.04), State=3.01 (+-0.01), Tradition=3.2 (+-0.08), Transgression=2.94 (+-0.03), Utilitarian=2.73 (+-0.07). For the full distribution, see Figure S3. Given the coding procedures employed, and similar to prior research, scores tend to cluster around 3. The averages show the overall profile of cultural styles within Toronto. Scenes there tend to have a strong emphasis on ethnic identity, tradition, and charisma. Yelp categories tend to include fewer venues geared toward activities like research, engineering, or industrial production, and hence scores for rationalism or utilitarianism are somewhat lower. It is important to place these values in context. For example, the national Canadian average for transgression based on the prior work noted above is below 2.8 -- this is an intuitive result, not only based on Canadian reputation but also in that transgressiveness almost by definition will dominate an entire scene relatively rarely. Thus, the fact that Toronto has a score around 3 (both with our mapping to Yelp and the previous scores based on yellow pages directories) indicates a relatively high level of transgressiveness, compared to the rest of the country. This mirrors US cities, in which dimensions like self-expression, glamour, corporateness, charisma, and utilitarianism are relatively higher in major cities, such as New York, Chicago, and Los Angeles [13]. Similarly, while the traditionalism score within Toronto is higher than many other dimensions, it is likely lower than traditionalism in a small time, and the lower rationalism scores might well be higher than the typical Canadian score. Overall Canadian patterns are mapped and probed more deeply in [14].

In the present context, of methodological importance is to note the typical range of values for these measures of local culture -- see Figure S3. This implies that improvements of RMSE on the order of 0.01–0.08 represent significant increases in model performance – see standard deviations above. Past

work has demonstrated that differences in scenes dimensions make a considerable difference for local outcomes such as changes in college graduates, age groups, and political orientations [15].

Table S2 summarizes the conceptual meaning of the cultural dimensions. It also includes some illustrative samplings of venues that are used as indicators for each of these dimensions.

| Cultural Dimension | Example | Sample Amenity Indicators |
|---|---|---|
| Glamour | Standing on the red carpet at Cannes gazing at the stars going by | Fashion shows & designers; designer clothes & accessories; beauty salons; nail salons; motion picture & video exhibition; motion picture & sound recording studios; agents, managers for artists & other public figures; film festivals; nightclubs; jewelry stores; casinos |
| Formal | Going to the opera in a gown or white tie and tails | Formal wear & costume rental; opera companies; fine dining; private clubs; dance companies; nightclubs; golf courses & country clubs; theater companies & dinner theater; religious organizations; offices of lawyers; professional organizations |
| Transgression | Watching a performance artist pierce his skin | Body piercing studios; tattoo parlors; adult entertainment: nightclubs; adult entertainment: comedy and dance clubs; leather clothing stores; skateboard parks; casinos; beer, wine & liquor stores; gambling industries; hemp shops |
| Neighbourliness | Attending a performance by the community orchestra | Bed & breakfast inns; civic & social organizations; religious organizations; golf courses & country clubs; sports teams & clubs; playgrounds; elementary & secondary schools; fruit & vegetable markets; coffee houses; pubs; baked goods stores |
| Exhibitionism | Watching weightlifters at Muscle Beach | Adult entertainment: night clubs; fashion shows & designers; body piercing; tattoo studios; health clubs; fashion shows & designers; beauty salons; nail salons; discotheques |
| Local | Listening to the blues in the Checkerboard Lounge, landmark of the Chicago blues | Bed & breakfast inns; historical sites; fishing lakes & ponds; marinas; book dealers: used & rare; antique dealers; scenic & sightseeing services; nature parks & other similar institutions; spectator sports; sports teams & clubs; microbreweries; fruit & vegetable markets; meat markets |
| Ethnic | Recognizing the twang of Appalachia in the Stanley Bros.' Voices | Ethnic Restaurants (approximately 40 cuisines); Ethnic Music; Ethnic Dance; Folk Arts; Cultural and Ethnic Awareness Programmes; Language Schools; Gospel Singing Groups; Martial Arts Instruction |
| Corporate | Reviling a Britney Spears show because she is a corporate creation | Marketing research; management consulting services; warehouse clubs & superstores; designer clothes & accessories; fast food restaurants; business & secretarial schools; department stores; convention & trade shows; public relations agencies; spectator sports; amusement & theme parks; advertising & related services |
| State | Visiting the Gettysburg Battlefield | Political organizations; embassies & delegations; historical sites; American restaurants |
| Rational | Revelling in the cosmic scope of human reason at a planetarium | R&D in physical, engineering & life sciences; scientific R&D services; colleges, universities & professional schools; planeteria; aquariums; human rights organizations; management, scientific & technical consulting; exam preparation & tutoring; libraries & archives; computer training; offices of lawyers |
| Traditional | Sharing in the stability and assurance of hearing Mozart performed in the Vienna State Opera as you believe it was earlier | Genealogy societies; historical sites; opera companies; antique dealers; fine arts schools; libraries & archives; family restaurants; family clothing stores; religious organizations; dance companies; museums; etiquette schools |
| Utilitarian | Attending a benefit concert because it contributes to positive outcomes or savouring the value of efficient production at a museum of industry | Fast food restaurants; technical & trade schools; warehouse clubs & superstores; business & secretarial schools; management consulting services; convenience stores; business associations; junior colleges; computer systems design; database & directory publishers; exam preparation & tutoring |
| Egalitarian | Enjoying the democratic implications of a crafts fair | Human rights organizations; Salvation Army; public libraries: elementary & secondary schools (public); environment & wildlife organizations; junior colleges; services for elderly & disabled persons; social advocacy organizations; individual & family services; religious organizations |
| Self-expressive | Enjoying hearing a jazz musician play something that could only be improvised spontaneously at that particular moment | Dance companies; fashion shows/designers; yoga studios; art dealers; comedy clubs; body piercing; tattoo parlors; recorded music stores; vintage & used clothing; custom printed t-shirts; music festivals: fine arts schools; graphic design services; independent artists, writers & performers; musical groups & artists; performing arts companies; sound recording industries; hobby, toy & game stores; interior design services; karaoke clubs |
| Charismatic | Watching a Chicago Bulls game because of the charismatic aura of Michael Jordan rather than because one is a Chicagoan | Designer clothes & accessories; fashion shows/designers; motion picture & video exhibition; art dealers; dance companies; historical sites; political organizations; capitols; motion picture & sound recording industries; musical groups & artists; performing arts companies; promoters of entertainment events; spectator sports; fine arts schools; sports bars; sound recording studios |

**Table S2.** Summarization of the conceptual meaning of the cultural dimensions and illustrative examples of the venues that may indicate them.

Figure S3 presents the cultural dimensions distributions for all FSAs in all years in Toronto.

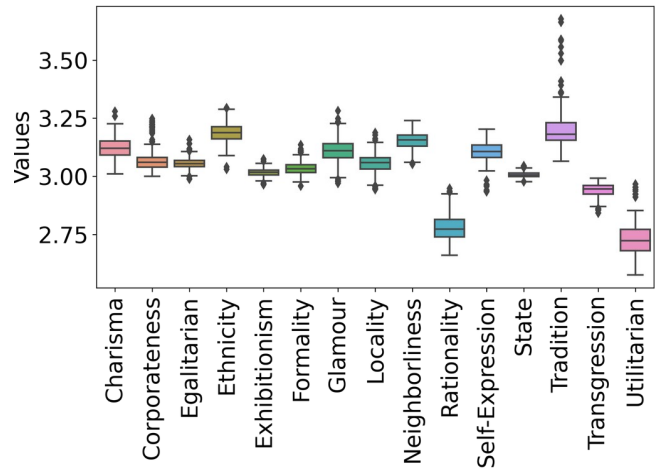

**Figure S3.** Cultural dimensions distributions for all FSAs in all years in Toronto.

## Extra Information on the Prediction Model

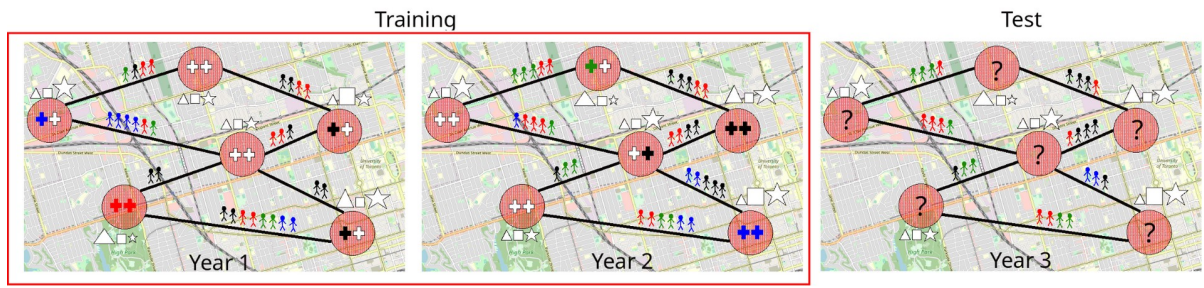

**Figure S4.** Our model explores mobility networks. In each evaluated scenario, we consider only one type of scenario (i.e., network) among the eight defined in Table 3.6. The figure illustrates the scenario "Area info node + mobility + profile." All networks have information on urban cultural dimensions in the areas (nodes). Our goal is to predict the urban cultural dimensions. If we envision predicting urban cultural dimensions for year y, we use all information on previous years available in our dataset to train our model (See Section Evaluation for details). [Best in colour]

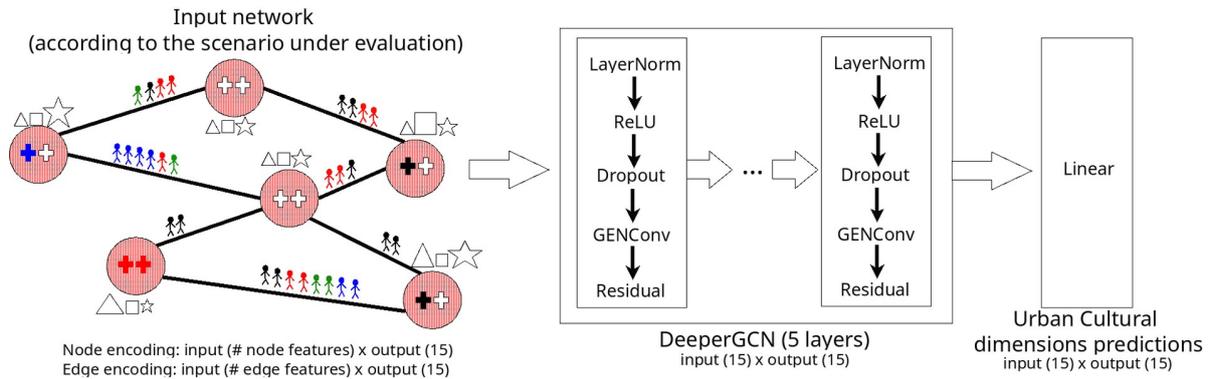

**Figure S5.** The GNN design. Edge and Node features of the input network are encoded by linear transformations. Multiple DeeperGCN layers then convolve the resulting encoded network. One DeeperGCN layer consists of a batch normalization layer (LayerNorm), an activation function (ReLU), one dropout layer, one GENConv layer, and a residual connection layer – details can be found in Li et al. [94]. A final linear layer maps the encoded information onto urban cultural dimensions predictions. [Best in colour]

**Census Files**

Figure S6 shows the differences between census variables explored from the two censuses considered, where the results represent values from the 2016 census minus the 2011 census. As we can see, most variables changed considerably during the period analyzed. Even the variables that changed a bit less, there are differences.

Calgary                                    Montreal

| | Edu | Rent | Vismin | Walk | Art | Income | Age |
|---|---|---|---|---|---|---|---|
| M1V | -0.012 | 1.4e+02 | 0.0087 | -0.0025 | 0.00064 | 9.2e+02 | 0.0034 |
| M2N | 0.023 | 2.4e+02 | 0.063 | 0.014 | 0.0037 | -1.8e+03 | 0.0022 |
| M4C | 0.045 | 1.8e+02 | 0.011 | -0.00036 | -0.0055 | 1.1e+03 | -0.0039 |
| M4E | 0.021 | 2.1e+02 | 0.029 | 0.011 | -0.0025 | 3.5e+03 | -0.0088 |
| M4J | 0.044 | 1.7e+02 | 0.027 | -0.0019 | -0.009 | 4.2e+03 | -0.021 |
| M4K | 0.044 | 1.7e+02 | 0.022 | 0.0081 | 0.0018 | 3.4e+03 | -0.0048 |
| M4L | 0.047 | 2.3e+02 | -0.018 | 0.015 | 0.011 | 4.1e+03 | -0.01 |
| M4M | 0.056 | 3.3e+02 | -0.045 | 0.027 | 0.0093 | 7.5e+03 | -0.0068 |
| M4P | 0.023 | 2.1e+02 | 0.04 | 0.017 | 0.002 | 8.4e+02 | -0.033 |
| M4R | 0.034 | 2.4e+02 | 0.061 | -0.006 | 0.016 | 4.1e+03 | -0.0074 |
| M4S | 0.032 | 2.3e+02 | 0.033 | 0.013 | 0.002 | 2.2e+03 | -0.031 |
| M4T | 0.036 | 2.3e+02 | 0.03 | 0.02 | 0.011 | 8.5e+03 | -0.0037 |
| M4V | -0.0041 | 2.1e+02 | 0.05 | 0.0013 | -0.0029 | 4.7e+03 | -0.0015 |
| M4W | 0.0078 | 2.3e+02 | 0.048 | 8.5e-05 | -0.0092 | 5.7e+03 | -0.0013 |
| M4X | 0.025 | 1.4e+02 | 0.029 | -0.0027 | -0.01 | 3.3e+03 | -0.013 |
| M4Y | 0.022 | 2.8e+02 | 0.056 | -0.004 | -0.0039 | 1.3e+03 | 0.02 |
| M5A | 0.062 | 2.8e+02 | -0.008 | 0.013 | -0.0044 | 4.2e+03 | 0.04 |
| M5B | 0.043 | 3.4e+02 | 0.058 | 0.025 | 0.0023 | 2.4e+03 | 0.027 |
| M5C | 0.035 | 96 | 0.039 | 0.11 | 0.016 | 6.4e+03 | -0.019 |
| M5E | 0.076 | 4.6e+02 | 0.022 | 0.02 | -0.01 | 1e+04 | 0.024 |
| M5G | 0.022 | 1.5e+02 | 0.046 | 0.083 | -0.0075 | 7.8e+02 | 0.015 |
| M5H | 0.13 | 7.6e+02 | 0.082 | 0.097 | -0.088 | 7.4e+03 | -0.073 |
| M5J | 0.032 | 2.1e+02 | 0.042 | 0.096 | -0.0068 | 5.9e+03 | 0.051 |
| M5M | 0.024 | 1.8e+02 | 0.052 | 0.0042 | -0.0021 | 7.6e+03 | -0.001 |
| M5P | 0.033 | 2.6e+02 | 0.034 | 0.00065 | -0.018 | 1e+04 | 0.0016 |
| M5R | 0.034 | 2.9e+02 | 0.029 | -0.0024 | -0.0081 | 7.4e+03 | -0.008 |
| M5S | 0.0069 | 3e+02 | 0.048 | 0.05 | -0.0083 | 2.9e+03 | 0.0017 |
| M5T | 0.029 | 1.8e+02 | -0.025 | 0.063 | 0.014 | 8.7e+02 | 0.015 |
| M5V | 0.049 | 2.8e+02 | 0.053 | 0.085 | 0.0009 | 3.6e+03 | 0.018 |
| M6A | 0.03 | 1.7e+02 | -0.011 | 0.0019 | 0.0028 | 1.5e+03 | 0.01 |
| M6C | 0.076 | 2e+02 | -0.019 | 0.00085 | 0.0072 | 4e+03 | -0.00098 |
| M6G | 0.031 | 2.5e+02 | -0.0037 | 0.028 | 0.0095 | 4.5e+03 | 0.0051 |
| M6H | 0.039 | 2.5e+02 | -0.0031 | 0.015 | -0.00048 | 4.4e+03 | 0.02 |
| M6J | 0.069 | 3.6e+02 | 0.021 | 0.033 | 0.011 | 6.8e+03 | 0.043 |
| M6K | 0.075 | 3.2e+02 | -0.0099 | 0.033 | 0.0071 | 6.9e+03 | 0.068 |
| M6P | 0.079 | 1.8e+02 | 0.003 | 0.003 | 0.015 | 5.8e+03 | -0.0096 |
| M6R | 0.041 | 2.4e+02 | -0.00045 | -0.0069 | 0.011 | 4.5e+03 | 0.0049 |
| M6S | 0.044 | 2.6e+02 | 0.018 | -0.0033 | 0.0037 | 7.8e+03 | 0.0041 |

Toronto

**Figure S6.** Differences between census variables explored from the two censuses considered (year 2016 minus 2011). Variables: Edu = percent with a BA or higher; Rent = average rent; Vismin = percent classified as ``visible minorities''; Walk = percent who walk to work; Art = percent who work in arts, entertainment, and culture; income = median income; Age = percent age 20-34. Colour legend: Blue = negative values, Gray = 0; Red = positive values.

**Extra Information on the Results**

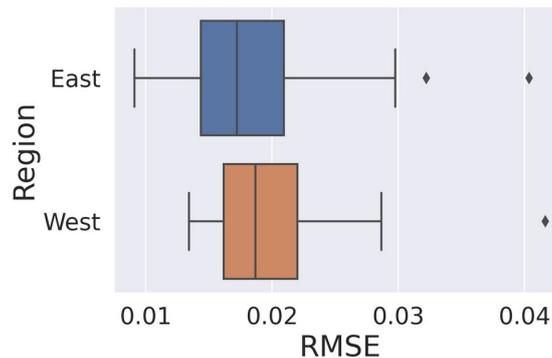

**Figure S7.** Prediction errors (RMSE) for FSA in west and east regions in Toronto, considering the scenario Area Info.

**References**

[1] Silver D, Clark TN and Navarro Yanez CJ. Scenes: Social context in an age of contingency. Social forces 2010; 88(5): 2293–2324.

[2] Silver D and Miller D. Contextualizing the artistic dividend. Journal of Urban Affairs 2013; 35(5): 591–606.

[3] Clark TN and Sawyer S. Villes créatives ou voisinages dynamiques? développement métropolitain et ambiances urbaines. L'Observatoire 2010; (1): 44–49.

[4] Mateos-Mora C, Navarro-Yañez CJ and Rodríguez-García MJ. A guide for the analysis of cultural scenes: a measurement proposal and its validation for the spanish case. Cultural Trends 2022; 31(4): 354–371

[5] Wu J, Zheng H, Wang T et al. Bohemian cultural scenes and creative development of chinese cities: An analysis of 65 cities using cultural amenity data. Sustainability 2021; 13(9): 5260

[6] Klekotko M et al. Mathematics of culture, or how to count the cultural character of a city? ASK Research & Methods 2021; (30): 81–104

[7] Jang W. Urban 'scenes' and local development: The case of seoul. Journal of Social Sciences 2012; 14: 1–23.

[8] Silver DA and Clark TN. Scenescapes: How qualities of place shape social life. University of Chicago Press, 2016.

[9] Silver D, Clark TN and Navarro Yanez CJ. Scenes: Social context in an age of contingency. Social forces 2010; 88(5): 2293–2324.

[10] Reimers N and Gurevych I. Sentence-bert: Sentence embeddings using siamese bert-networks. arXiv preprint arXiv:190810084 2019; .

[11] Gubert F and Silva T. Google places enricher: A tool that makes it easy to get and enrich google places api data. In Anais Estendidos do XXVIII Simpósio Brasileiro de Sistemas Multimídia e Web. Porto Alegre, RS, Brasil: SBC, pp. 91–94. DOI:10.5753/webmedia estendido.2022.227245.

[12] Gubert F. Google places enricher, 2023. URL https://github.com/FerGubert/google_places_enricher. Acessed March 4, 2023.

[13] Silver D. The american scenescape: Amenities, scenes and the qualities of local life. Cambridge Journal of Regions, Economy and Society 2012; 5(1): 97–114.

[14] Silver D. Some Scenes of Urban Life. SAGE, 2017.

[15] Miller DL and Silver D. Cultural scenes and contextual effects on political attitudes. European journal of cultural and political sociology 2015; 2(3-4): 241–266.



\end{document}